\pdfoutput=1

\documentclass[11pt]{article}

\usepackage{EMNLP2023}

\usepackage{times}
\usepackage{latexsym}

\usepackage[T1]{fontenc}

\usepackage[utf8]{inputenc}

\usepackage{microtype}

\usepackage{inconsolata}
\usepackage{url}
\usepackage{amssymb}
\usepackage{graphicx}
\usepackage{booktabs}
\usepackage{CJKutf8}
\usepackage{graphicx}
\usepackage{subfigure}
\usepackage{float}
\usepackage{amsmath}
\usepackage{amssymb}
\usepackage{multirow}
\usepackage{booktabs}
\usepackage{url}
\usepackage{array}
\usepackage{enumitem}
\usepackage{algorithm}
\usepackage{algorithmic}
\usepackage{pifont}
\usepackage{bm}
\usepackage{lipsum}
\usepackage{xcolor}
\usepackage{makecell}
\usepackage{CJKutf8}
\usepackage{tikz}
\usepackage[edges]{forest}
\definecolor{hidden-draw}{RGB}{20,68,106}
\definecolor{hidden-pink}{RGB}{255,245,247}

\newcommand{\xmark}{\ding{55}}%
\title{Aligning Large Language Models with Human: A Survey}

\author{\textbf{Yufei Wang}, \textbf{Wanjun Zhong}, \textbf{Liangyou Li}, \textbf{Fei Mi},  \textbf{Xingshan Zeng}, \textbf{Wenyong Huang} \\ \textbf{Lifeng Shang}, \textbf{Xin Jiang}, \textbf{Qun Liu} \\
Huawei Noah's Ark Lab
\\
\{wangyufei44,zhongwanjun1,liliangyou,mifei2,zeng.xingshan,wenyong.huang\}@huawei.com \\
\{Shang.Lifeng,Jiang.Xin,qun.liu\}@huawei.com
}

\begin{document}
\maketitle
\begin{abstract}

Large Language Models (LLMs) trained on extensive textual corpora have emerged as leading solutions for a broad array of Natural Language Processing (NLP) tasks. 
Despite their notable performance, these models are prone to certain limitations such as misunderstanding human instructions, generating potentially biased content, or factually incorrect (hallucinated) information. Hence, aligning LLMs with human expectations has become an active area of interest within the research community.
This survey presents a comprehensive overview of these alignment technologies, including the following aspects. 
(1) \textbf{Data collection}:  the methods for effectively collecting high-quality instructions for LLM alignment, including the use of NLP benchmarks, human annotations, and leveraging strong LLMs.
(2) \textbf{Training methodologies}: a detailed review of the prevailing training methods employed for LLM alignment. Our exploration encompasses Supervised Fine-tuning, both Online and Offline human preference training, along with parameter-efficient training mechanisms.
(3) \textbf{Model Evaluation}: the methods for evaluating the effectiveness of these human-aligned LLMs, presenting a multifaceted approach towards their assessment.
In conclusion, we collate and distill our findings, shedding light on several promising future research avenues in the field. This survey, therefore, serves as a valuable resource for anyone invested in understanding and advancing the alignment of LLMs to better suit human-oriented tasks and expectations. An associated GitHub link collecting the latest papers is available at~\url{https://github.com/GaryYufei/AlignLLMHumanSurvey}.
\end{abstract}

\section{Introduction}
Foundational Large Language Models (LLMs) such as GPT-3 are pre-trained on a vast textual corpus with objectives to predict subsequent tokens. This process equips LLMs with world knowledge, facilitating the generation of coherent and fluent text in response to various inputs. 
Despite these strengths, foundational LLMs are not always adept at interpreting a wide range of instructions and can produce outputs that deviate from human expectations. Additionally, these models may produce biased content or invent (hallucinated) facts, which can limit their practical usefulness.

Therefore, recent NLP research efforts focus on empowering LLMs to understand instructions and to align with human  expectations. 
Early methods for training LLMs to follow instructions primarily use task instruction sets, which are compiled by combining manually crafted task instruction templates with instances from standard NLP tasks.
However, such approaches often fall short of capturing the intricacies of practical user instructions, as these instructions tend to originate from artificial NLP tasks designed to test specific aspects of machine capabilities. Real-world user instructions, on the other hand, are significantly more diverse and complex.
As a result, OpenAI explored Supervised Fine-Tuning (SFT) of LLMs using instructions annotated by a diverse group of human users. Models developed through this process, such as InstructGPT~\cite{DBLP:conf/nips/Ouyang0JAWMZASR22} and ChatGPT~\footnote{\url{https://chat.openai.com/}}, have demonstrated a marked improvement in understanding human instructions and solving complex tasks.
To further enhance alignment, ~\citet{DBLP:conf/nips/Ouyang0JAWMZASR22} incorporate the Reinforcement Learning from Human Feedback (RLHF) approach, which involves learning from human preferences through a reward model trained with human-rated outputs.

\tikzstyle{my-box}=[
    rectangle,
    draw=hidden-draw,
    rounded corners,
    text opacity=1,
    minimum height=1.5em,
    minimum width=5em,
    inner sep=2pt,
    align=center,
    fill opacity=.5,
    line width=0.8pt,
]
\tikzstyle{leaf}=[my-box, minimum height=1.5em,
    fill=hidden-pink!80, text=black, align=left,font=\normalsize,
    inner xsep=2pt,
    inner ysep=4pt,
    line width=0.8pt,
]
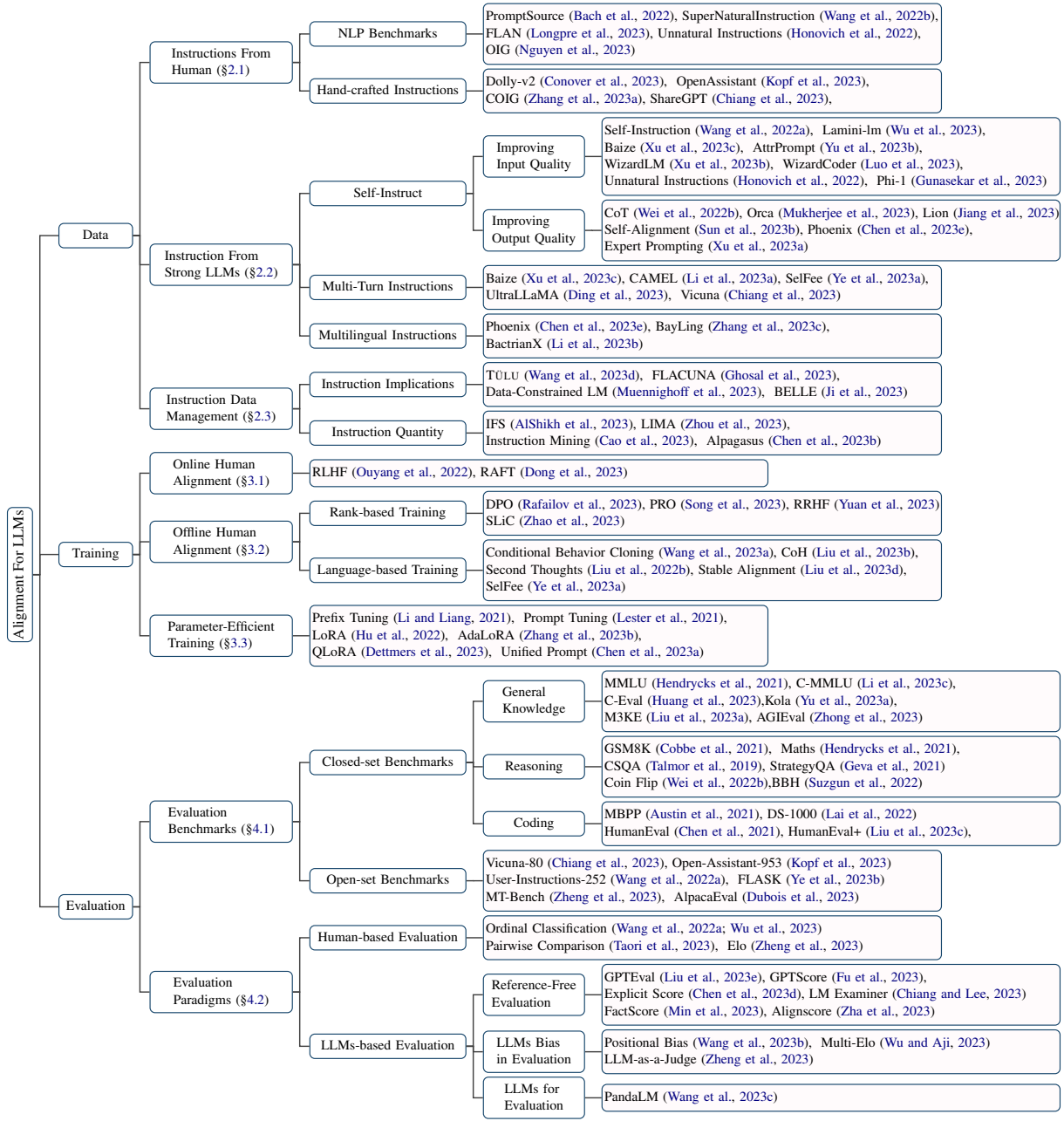
\begin{figure*}[t!]
    \centering
    \resizebox{\textwidth}{!}{
        \begin{forest}
            forked edges,
            for tree={
                grow=east,
                reversed=true,
                anchor=base west,
                parent anchor=east,
                child anchor=west,
                base=center,
                font=\large,
                rectangle,
                draw=hidden-draw,
                rounded corners,
                align=left,
                text centered,
                minimum width=4em,
                edge+={darkgray, line width=1pt},
                s sep=3pt,
                inner xsep=2pt,
                inner ysep=3pt,
                line width=0.8pt,
                ver/.style={rotate=90, child anchor=north, parent anchor=south, anchor=center},
            },
            where level=1{text width=5em,font=\normalsize,}{},
            where level=2{text width=10em,font=\normalsize,}{},
            where level=3{text width=11em,font=\normalsize,}{},
            where level=4{text width=7em,font=\normalsize,}{},
            [
                Alignment For LLMs, ver
                [
                    Data 
                    [
                        Instructions From \\ Human  (\S \ref{instructionfromhuman})
                        [
                            NLP Benchmarks
                            [                            
                                PromptSource~\cite{bach-etal-2022-promptsource}{, }SuperNaturalInstruction~\cite{wang-etal-2022-super}{, } \\ FLAN~\cite{longpre2023flan}{, }Unnatural Instructions~\cite{DBLP:journals/corr/abs-2212-09689}{, } \\
                                OIG~\cite{OIG}
                                , leaf, text width=33em
                            ]
                        ]
                        [
                            Hand-crafted Instructions
                            [
                                Dolly-v2~\cite{DatabricksBlog2023DollyV2}{, } OpenAssistant~\cite{Kopf2023OpenAssistantC}{, } \\
                                COIG~\cite{Zhang2023ChineseOI}{,} ShareGPT~\cite{vicuna2023}{,} 
                                , leaf, text width=33em
                            ]
                        ]
                    ]
                    [
                        Instruction From \\ Strong LLMs (\S \ref{llminstruction})
                        [
                            Self-Instruct
                            [
                                Improving \\ Input Quality
                                [
                                    Self-Instruction~\cite{DBLP:journals/corr/abs-2212-10560}{, } Lamini-lm~\cite{DBLP:journals/corr/abs-2304-1440}{, } \\ Baize~\cite{DBLP:journals/corr/abs-2304-01196}{, } AttrPrompt~\cite{yu2023large}{, }\\ WizardLM~\cite{xu2023wizardlm}{, } WizardCoder~\cite{luo2023wizardcoder}{, }\\
                                    Unnatural Instructions~\cite{DBLP:journals/corr/abs-2212-09689}{, } Phi-1~\cite{gunasekar2023textbooks}
                                    , leaf, text width=33em
                                ]
                            ]
                            [
                                Improving \\Output Quality
                                [
                                    CoT~\cite{wei2022chain}{, }Orca~\cite{mukherjee2023orca}{, }Lion~\cite{Jiang2023LionAD}\\
                                    Self-Alignment~\cite{Sun2023PrincipleDrivenSO}{, }Phoenix~\cite{DBLP:journals/corr/abs-2304-10453}{, }\\
                                    Expert Prompting~\cite{xu2023expertprompting}
                                    , leaf, text width=33em
                                ]
                            ]
                        ]
                        [
                            Multi-Turn Instructions
                            [
                                Baize~\cite{DBLP:journals/corr/abs-2304-01196}{, }CAMEL~\cite{DBLP:journals/corr/abs-2303-17760}{, }SelFee~\cite{selfee2023}{, }\\UltraLLaMA~\cite{ding2023enhancing}{, } Vicuna~\cite{vicuna2023} 
                            , leaf, text width=33em
                            ]
                        ]
                        [
                            Multilingual Instructions
                            [
                                Phoenix~\cite{DBLP:journals/corr/abs-2304-10453}{, }BayLing~\cite{Zhang2023BayLingBC}{, }\\ BactrianX~\cite{bactrian}
                            , leaf, text width=33em
                            ]
                        ]
                    ]
                    [
                        Instruction Data \\ Management (\S \ref{datamanagement})
                        [
                            Instruction Implications
                            [
                                \textsc{T\"ulu}~\cite{wang2023far}{, } FLACUNA~\cite{ghosal2023flacuna}{, } \\
                                Data-Constrained LM~\cite{muennighoff2023scaling}{, }  BELLE~\cite{DBLP:journals/corr/abs-2304-07854}
                                , leaf, text width=33em
                            ]
                        ]
                        [
                            Instruction Quantity
                            [
                                IFS~\cite{alshikh2023becoming}{, }LIMA~\cite{zhou2023lima}{, } \\
                                Instruction Mining~\cite{cao2023instruction}{, }
                                Alpagasus~\cite{chen2023alpagasus}
                                , leaf, text width=33em
                            ]
                        ]
                    ]
                ]
                [
                    Training
                    [
                        Online Human \\ Alignment (\S \ref{onlinetraining})
                        [
                            RLHF~\cite{DBLP:conf/nips/Ouyang0JAWMZASR22}{, }RAFT~\cite{dong2023raft}
                            , leaf, text width=33em
                        ]
                    ]
                    [
                        Offline Human \\ Alignment (\S \ref{offlinetraining})
                        [
                            Rank-based Training
                            [
                                DPO~\cite{rafailov2023direct}{,} PRO~\cite{song2023preference}{,} RRHF~\cite{yuan2023rrhf} \\
                                SLiC~\cite{zhao2023calibrating} 
                                , leaf, text width=33em
                            ]
                        ]
                        [
                            Language-based Training
                            [
                                Conditional Behavior Cloning~\cite{openchat}{, }CoH~\cite{liu2023languages}{, } \\
                                Second Thoughts~\cite{liu2022second}{, }Stable Alignment~\cite{liu2023training}{, }\\ SelFee~\cite{selfee2023}
                                , leaf, text width=33em
                            ]
                        ]
                    ]
                    [
                        Parameter-Efficient \\ Training (\S \ref{parametereffective})
                        [
                            Prefix Tuning~\cite{li-liang-2021-prefix}{, } Prompt Tuning~\cite{lester-etal-2021-power}{, } \\
                            LoRA~\cite{hu2022lora}{, } AdaLoRA~\cite{zhang2023adaptive}{, } \\
                            QLoRA~\cite{dettmers2023qlora}{, } Unified Prompt~\cite{chen2023parameterefficient}
                            , leaf, text width=33em
                        ]
                    ]
                ]
                [
                    Evaluation
                    [
                        Evaluation \\ Benchmarks (\S \ref{evalbenchmark})
                        [
                            Closed-set Benchmarks
                            [
                                General \\ Knowledge
                                [MMLU~\cite{hendrycks2021measuring}{,} C-MMLU~\cite{li2023cmmlu}{,}\\C-Eval~\cite{huang2023ceval}{,}Kola~\cite{yu2023kola}{,}\\ M3KE~\cite{liu2023m3ke}{,} AGIEval~\cite{zhong2023agieval} 
                                , leaf, text width=33em]
                            ]
                            [
                                Reasoning
                                [GSM8K~\cite{cobbe2021training}{, } Maths~\cite{hendrycks2021measuring}{, }\\CSQA~\cite{talmor-etal-2019-commonsenseqa}{,} StrategyQA~\cite{geva-etal-2021-aristotle}\\Coin Flip~\cite{wei2022chain}{,}BBH~\cite{suzgun2022challenging}
                                , leaf, text width=33em]
                            ]
                            [
                                Coding
                                [MBPP~\cite{austin2021program}{, }DS-1000~\cite{Lai2022DS1000}\\HumanEval~\cite{chen2021evaluating}{, }HumanEval+~\cite{liu2023your}{, }
                                , leaf, text width=33em]
                            ]
                        ]
                        [
                            Open-set Benchmarks
                            [
                                Vicuna-80~\cite{vicuna2023}{, }Open-Assistant-953~\cite{Kopf2023OpenAssistantC}\\
                                User-Instructions-252~\cite{DBLP:journals/corr/abs-2212-10560}{, } FLASK~\cite{Ye2023FLASKFL} \\
                                MT-Bench~\cite{zheng2023judging}{, }  AlpacaEval~\cite{dubois2023alpacafarm}
                                , leaf, text width=33em
                            ]
                        ]
                    ]
                    [
                        Evaluation \\ Paradigms (\S \ref{evalparadigm})
                        [
                            Human-based Evaluation 
                            [
                                Ordinal Classification~\cite{DBLP:journals/corr/abs-2212-10560,DBLP:journals/corr/abs-2304-1440} \\
                                Pairwise Comparison~\cite{alpaca}{, }
                                 Elo~\cite{zheng2023judging}
                                , leaf, text width=33em
                            ]
                        ]
                        [
                            LLMs-based Evaluation 
                            [
                                Reference-Free \\ Evaluation
                                [
                                GPTEval~\cite{liu2023gpteval}{, }GPTScore~\cite{fu2023gptscore}{,}\\Explicit Score~\cite{chen2023exploring}{, }LM Examiner~\cite{chiang2023can}\\
                                FactScore~\cite{min2023factscore}{, }Alignscore~\cite{zha2023alignscore}
                                , leaf, text width=33em
                                ]
                            ]
                            [
                                LLMs Bias \\ in Evaluation 
                                [Positional Bias~\cite{wang2023large}{, } 
                            Multi-Elo~\cite{Wu2023StyleOS} \\
                            LLM-as-a-Judge~\cite{zheng2023judging}
                            , leaf, text width=33em]
                            ]
                            [
                                LLMs for \\ Evaluation
                                [PandaLM~\cite{wang2023pandalm}
                                , leaf, text width=33em]   
                            ]
                        ]
                    ]
                ]
            ]
        \end{forest}
    }
    \caption{Taxonomy of research in aligning Large Language Models (LLMs) with human that consists of alignment data, training strategy, and evaluation methods.}
    \label{fig:sftframework}
\end{figure*}

There are challenges in alignment processes and the subsequent evaluation:
(a) Collecting high-quality data for both SFT and RLHF stages can be costly and time-consuming.
(b) The training strategies need to be optimized as SFT training is resource-consuming, and reinforcement learning in RLHF often lacks stability.
(c) Evaluating LLMs comprehensively is challenging, as limited NLP benchmarks may not fully reveal the multifaceted capabilities of LLMs.

To address these limitations, extensive research efforts have been devoted. In Figure~\ref{fig:sftframework}, we provide a summary of these multi-aspect approaches.
For aspect (a), the focus is on effectively collecting large-scale, high-quality data for LLM alignment training. Researchers propose leveraging the power of existing NLP benchmarks, human annotators, and state-of-the-art LLMs (e.g., ChatGPT and GPT-4) to generate training instructions.
To tackle aspect (b), solutions involve optimizing the training methods for better efficiency and stability in incorporating human preferences. 
Parameter-efficient training methods have been proposed to reduce computation burden and improve efficiency in LLM alignment. 
Additionally, some researchers consider human preference as ranking-based training signals or replace scalar rewards with language-based feedback to enhance training stability and performance.
Regarding aspect (c), various human-centric LLM evaluation benchmarks and automatic evaluation protocols (e.g., LLMs for evaluation) have been proposed to obtain a comprehensive evaluation of aligned LLMs.

In this survey, we aim to provide a comprehensive overview of alignment technologies for large language models. 
In Section~\ref{instructioncollecting}, we summarize various methods in effective high-quality data collection. Section~\ref{training} focuses on popular training methods to incorporate human preference data into LLMs. The evaluation benchmarks and automatic protocols for instruction-following LLMs are discussed in Section~\ref{eval}. 
By collating and distilling our findings, we shed light on several promising future research avenues in Section~\ref{challengesdirection}.
Through this survey, we aim to provide an overview of the current state of LLM alignment, enabling researchers and practitioners to navigate the complexities of aligning LLMs with human values and expectations.

\section{Alignment Data Collection}
\label{instructioncollecting}
Aligning LLMs with human expectations necessitates the collection of high-quality training data that authentically reflects human needs and expectations. For the purposes of this survey, we conceptualize an instruction as $I_k = (x_k, y_k)$, where $x_k$ denotes the instruction input and $y_k$ denotes the corresponding response. This data can be derived from an array of sources, encompassing both human-generated instructions and those generated by strong LLMs.
In this section, we summarize these methods of instruction generation and effective strategies for constructing a composite of diverse training instructions.


\subsection{Instructions from Human}
\label{instructionfromhuman}
Human-provided instructions mainly originate from two main sources: pre-existing human-annotated NLP benchmarks and meticulously hand-crafted instructions. 

\subsubsection{NLP Benchmarks}
\begin{figure}[!ht]
    \centering
    \includegraphics[width=\columnwidth]{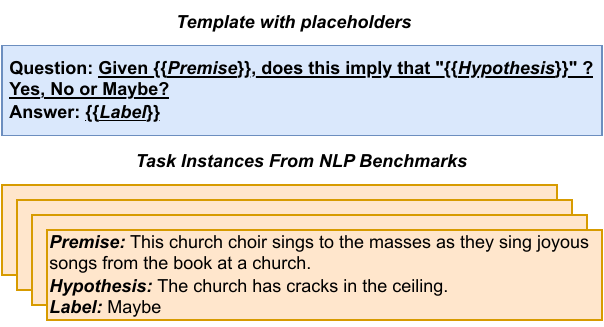}
    \caption{An Example of Instruction from a Natural Language Inference (NLI) benchmark.}
    \label{fig:nlpinstruction}
\end{figure}

An intuitive starting point for data collection involves adapting existing NLP benchmarks into natural language instructions. For instance, Figure~\ref{fig:nlpinstruction} offers an example drawn from the Natural Language Inference task. Works such as PromptSource~\cite{bach-etal-2022-promptsource}, FLAN~\cite{wei2022finetuned,longpre2023flan}, and SuperNaturalInstruction~\cite{wang-etal-2022-super,mishra-etal-2022-cross} are at the forefront of this approach. 
These benchmarks represent a substantial array of \emph{diverse and heterogeneous} NLP tasks, such as dialogue, reasoning tasks and coding tasks, unified under the framework of language instructions. In each NLP benchmark, they engage annotators to craft several natural language templates that smoothly integrate all input data into a sequential text. The objective is to enhance LLMs' capability for multi-task learning across training tasks and foster generalization for unseen tasks. OIG~\cite{OIG} also combines instructions from FLAN-like NLP benchmarks with other types of open-ended instructions, such as how-to, maths and coding instructions. Concurrently, \citet{DBLP:journals/corr/abs-2212-09689} put forth the concept of \emph{Unnatural Instructions}, utilizing LLMs to generate new templates or instances bearing resemblance to the original instructions but with notable variances. Interestingly, the authors discovered that \emph{text-davinci-002} outperforms GPT-3 in responding to these generated instructions, given that GPT-3 often devolved into repetitive or tangential outputs after providing the correct answer. This model of instruction creation is highly scalable and can yield millions of instructions effectively. Further, \citet{wang2023far} demonstrated that FLAN-style instructions considerably enhanced the reasoning capabilities of aligned LLMs.

\subsubsection{Hand-crafted Instructions}
Constructing instructions from NLP benchmarks could be effective and painless. However, as many NLP datasets focus on a small and specific skill set, which means the resultant instructions are also relatively narrow in scope. Consequently, they may fall short in catering to the complex needs of real-world applications, such as engaging in dynamic human conversation.

To combat the above issues, it is possible to construct instructions via intentional manual annotations. How to effectively design a human-in-the-loop annotation framework becomes the key issue. The Databricks company collects a 15k crowd-sourcing instruction dataset \emph{databricks-dolly-15k}~\cite{DatabricksBlog2023DollyV2} from its employees. 
Those people are instructed to create prompt / response pairs in each of eight different instruction categories, including the seven outlined in~\citet{DBLP:conf/nips/Ouyang0JAWMZASR22}, as well as an open-ended free-form category. Importantly, they are \emph{explicitly} instructed not to use external web information, as well as outputs from generative AI systems.
~\citet{Kopf2023OpenAssistantC} construct the \emph{OpenAssistant} corpus with over 10,000 dialogues using more than 13,000 international annotators. The annotation process includes a) writing initial prompts for dialogue; b) replying as an assistant or user; c) ranking dialogue quality to explicitly provide human preferences. As a result, this corpus can be used for SFT and human preference alignment training for LLMs.
~\citet{Zhang2023ChineseOI} construct high-quality Chinese instructions from existing English instruction datasets. They first translate the English instructions into Chinese, then verify whether these translations are usable. Finally, they hire annotators to 
correct and re-organize the instructions into the {task description, input, output} format in the selected corpus. ShareGPT~\footnote{\url{https://sharegpt.com/}}, which is collected by~\citet{vicuna2023}, is an interesting exploration for crowd-sourcing human-written instructions. It is a website that encourages users to upload and share their interesting ChatGPT/GPT4 conversations. Such a mechanism can effectively collect a large number of diverse and human-written instructions that likely trigger high-quality ChatGPT/GPT4 responses. Popular online QA websites, such as Stack Overflow~\footnote{\url{https://stackoverflow.com/}}, Quora~\footnote{\url{https://www.quora.com/}} and Zhihu~\footnote{\url{https://www.zhihu.com/}}, and large user-generated content databases, such as Wikipedia~\footnote{\url{https://en.wikipedia.org/}}, are all reliable sources to provide high-quality human-written prompts for this purpose.Both ~\citet{ding2023enhancing} and~\citet{DBLP:journals/corr/abs-2304-01196} propose to use these resources as the seed instructions to prompt GPT-3.5 to generate high-quality synthetic multi-turn dialogues.

\subsection{Instructions From Strong LLMs}
\label{llminstruction}
With the emergence of strong closed-source LLMs (e.g., ChatGPT/GPT4), 
it is also feasible to automate the collection process to obtain various types of synthetic instructions (e.g., single-turn, multi-turn, and multilingual instructions) by providing appropriate prompts to these LLMs. The main challenge is how to effectively prompt LLMs to generate diverse and high-quality instructions.   

\begin{figure}[!ht]
    \centering
    \includegraphics[width=\columnwidth]{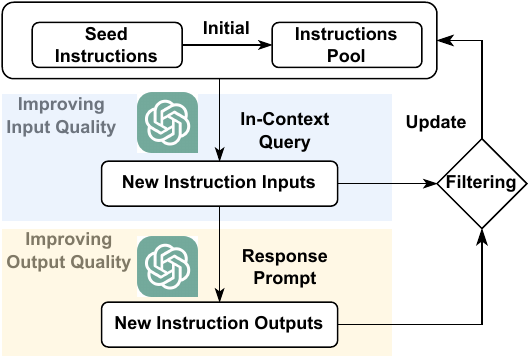}
    \caption{The overview of self-instruction. Starting from instructions in the pool, self-instruction leverages LLMs to produce new instructions via in-context learning. After filtering, LLMs are then prompted to respond to the remaining instructions. The full instructions are then added to the pool. Research efforts have been devoted to 1) Improving instruction input quality, and 2) Improving instruction output quality.}
    \label{fig:selfinstruct}
\end{figure}

\subsubsection{Self-Instruction}

\emph{Self-Instruct}~\cite{DBLP:journals/corr/abs-2212-10560} were among the pioneers to automate the instruction collection process.
It employed the in-context learning capability of ChatGPT to generate large-scale instructions from a pre-defined set of human-annotated instructions covering diverse topics and task types, as illustrated in Figure~\ref{fig:selfinstruct}. 
The automatically generated instructions are followed by a quality control filtering process, and this iterative process continues until the desired data volume has been achieved. Interestingly, the researchers discovered that GPT-3~\cite{NEURIPS2020_1457c0d6}, fine-tuned with these instructions, performed better than models fine-tuned using instructions derived from NLP benchmarks SuperNI benchmark~\cite{wang-etal-2022-super} and \emph{User-Oriented Instructions}, as discussed in Section~\ref{instructionfromhuman}).
Several follow-up attempts, such as Aplaca~\cite{alpaca} and its variants~\cite{chinese-llama-alpaca} follow this 
\emph{Self-Instruct} framework.
More subsequent research efforts w.r.t. enhancing instruction diversity, quality, and complexity will be elaborated as follows.

\paragraph{Improving Input Quality}
One limitation of the synthetic instructions from strong LLMs often suffer from diversity issues. For example,~\citet{jentzsch2023chatgpt} find that when prompting to generate jokes, ChatGPT only produces 25 unique joke patterns in thousands of samples. 
To improve the instruction input diversity,~\citet{DBLP:journals/corr/abs-2212-10560} propose different input and output generation strategies for different types of instructions. 
They first prompt ChatGPT to classify generated instruction into \emph{classification tasks} or \emph{non-classification tasks}. Then, they deploy output-first and input-first strategies for \emph{classification tasks} and \emph{non-classification tasks}, respectively.
Others propose to add various external information into the input prompts to enhance diversity and factuality, including Wikipedia Category Keywords~\cite{DBLP:journals/corr/abs-2304-1440}, user-generated questions on the Internet (e.g., Quora, StackOverflow)~\cite{DBLP:journals/corr/abs-2304-01196,gpt4all} and instructions from the SuperNaturalInstruction benchmark~\cite{DBLP:journals/corr/abs-2212-09689}.
\citet{yu2023large} also shows that explicitly adding meta-information (e.g., length, topics, style) into the data generation prompts can effectively remove the bias in the generated synthetic data and improve the diversity of those synthetic data.
Furthermore,~\citet{xu2023wizardlm} propose a novel \emph{Evol-Instruct} framework to obtain complex and difficult instructions gradually. 
Instead of using existing instructions to prompt LLMs to produce new instructions via \emph{in-context learning}, in \emph{Evol-Instruct}, there are five different manually-designed prompts to explicitly instruct LLMs to rewrite the existing simple instructions into complex ones using in-depth methods (i.e., including more information on particular topics) or in-Breadth methods (i.e, improving topics/information coverage). The resulting WizardLM model is ranked top in the MT-Bench~\cite{zheng2023judging} and AlpacaEval~\cite{dubois2023alpacafarm}.
\citet{luo2023wizardcoder} further expand this idea to produce complex code and programming instructions from the simple ones and propose the \emph{WizardCoder} model, which outperforms several strong commercial LLMs, e.g., Anthropic's Claude and Google's Bard.~\citet{gunasekar2023textbooks} propose to generate textbook-like instructions prompted with sufficient background knowledge to promote reasoning and basic algorithmic skills of LLMs. They find that the resulting 1.3B LLMs \emph{phi-1} successfully outperform various much larger LLMs, showing the importance of data quality.

\paragraph{Improving Output Quality}
Aside from the provision of high-quality instruction input, a critical requisite is to skillfully prompt LLMs to yield high-quality responses. The conventional method of enhancing response quality entails appending LLM prompts with additional conditions, encompassing the following facets.

\textbf{(1) Reasoning-Provoking Conditions:} ~\citet{wei2022chain} proposed the Chain-of-Thought (CoT) reasoning approach, which includes preconditions in the LLM prompts and  generation the intermediate reasoning processes for complex problems, thereby assisting LLMs in problem-solving. Inspired by CoT, ~\citet{mukherjee2023orca} developed the Orca model, which learns not only the superficial response text from LLMs, but also captures complex reasoning process signals. Specifically, they guided LLMs to respond to reasoning-intensive FLAN instructions with a series of predefined system prompts (e.g., ``think step-by-step and justify your response''), spurring LLMs (e.g., GPT4) to disclose their reasoning process information. Thanks to these advancements, the Orca model significantly outperformed several powerful open-sourced LLMs.

\textbf{(2) Hand-crafted Guiding Principles:} ~\citet{Sun2023PrincipleDrivenSO} introduced  \emph{self-alignment} framework that incorporates 16 manually devised principle rules into input prompts, thereby steering LLMs towards generating useful, ethical, and reliable responses. To augment the impact of these rules, they employed the Chain-of-Thoughts (CoT) technology~\cite{wei2022chain}, elucidating five examples to coach LLMs in discerning which rules to implement prior to generating actual response contents.

\textbf{(3) Role-playing Conditions:} ~\citet{DBLP:journals/corr/abs-2304-10453} devised a method to generate a set of role profiles using a blend of ChatGPT and manual efforts. They created seed instructions for each role profile and applied \emph{self-instruction} to the combination of role profiles and instructions to obtain nuanced responses from LLMs.~\citet{xu2023expertprompting} proposed a two-stage instruction response framework in which an expert profile is initially generated based on the instructions to be answered, followed by using both the expert profile and actual instructions to prompt LLMs for high-quality responses. 

\textbf{(4) Difficulty-monitoring Conditions:} 
~\citet{Jiang2023LionAD} proposed monitoring the quality of instruction response based on external LLM-based evaluations. They first fine-tune foundational LLMs with instruction data to obtain ``student LLMs''. Then, for each of training instruction, they gather responses from both teacher LLMs (e.g., ChatGPT) and student LLMs and prompted LLMs to conduct pairwise evaluation on the quality of both responses. Instructions are retained only when the student LLMs' response falls short of that from the teacher LLMs.

\subsubsection{Multi-turn Instructions}
In previous sections, we mainly focus on collecting synthetic single-turn instructions. However, LLMs well aligned with human should be capable to interact with users in a dialogue-based setting. To achieve this goal, some research efforts attempt to collect synthetic multi-turn instructions from strong LLMs.
When aligning LLaMA with human, Vicuna~\cite{vicuna2023} leverage instructions from ShareGPT which is website hosting interesting human-LLMs joint conversations. However, ShareGPT requires large volumes of users to upload their conversations.
~\citet{DBLP:journals/corr/abs-2304-01196} propose a novel Self-Chatting framework where questions from popular QA websites are used as the starting topics, then Chat-3.5 is prompted to chat with itself about this question in a four-turn dialogue.~\citet{DBLP:journals/corr/abs-2303-17760} propose \emph{CAMEL}, a ``role-playing'' framework where a human annotators first provide a topic, then LLMs are separately prompted to be ``AI Users'' and ``AI Assistants'' to discuss about this topic.~\citet{DBLP:journals/corr/abs-2304-07854} take a step further and prompt LLMs to first determine the conversation topic and then ask LLMs to chat with themselves to produce dialogue corpus.~\citet{selfee2023} propose a novel revision-based multi-turn dialogue corpus. Specifically, after instructions and initial responses, they further prompt LLMs to generate feedback and the revised version of responses if necessary. They use this dataset to train the \emph{SelFee} model and show that \emph{SelFee} can effectively improve its own answers when prompted to do so without any external guidance. The UltraLLaMA model~\cite{ding2023enhancing} leverages a wide range of real-world information, including (a) real-world knowledge from LLMs and Wikipedia; (b) various text creation tasks; (c) high-quality textual corpus, to produce initial questions and instructions that guide LLMs to generate diverse and high-quality multi-turn dialogues.

\subsubsection{Multilingual Instructions}
The above-generated instructions or dialogues are mostly based on English. To align LLMs with human who speak other languages, it is urgent and essential to expand the existing English resources into Multilingual ones. One straightforward idea is to translate instruction inputs and outputs into the target languages.
~\citet{DBLP:journals/corr/abs-2304-10453} propose two translation strategies: \emph{(a)} Post-answering which first translates the instruction inputs into the target language and then prompts strong LLMs to answer it. This could potentially preserve the specific culture patterns embedded in the target languages, but the output quality may be low as existing strong LLMs are often English-dominated; \emph{(b)} Post-translating which first prompts strong LLMs to respond the instructions in English, then translate both inputs and outputs. This approach could obtain high-quality output text, but lost the specific culture information.~\citet{bactrian} follow the \emph{Post-answering} strategy to construct instruction data for 52 popular languages using Google-Translate, then use these data to fine-tune LLaMA using the LoRA technology. An alternative solution is to mix several langauges in a multi-turn dialogue. BayLing~\cite{Zhang2023BayLingBC} introduces a set of multi-turn \emph{interactive translation} instructions to simultaneously improve multilingual and instruction-following ability for LLMs. Specifically, each multi-turn instruction is essentially a translation task where users first ask LLMs to translate a sentence to another language, then the users gradually add additional requirements (e.g., could you only use 10 words?). This process naturally connects different languages as well as human preferences with LLMs. We also summarize how to effectively adapt English-oriented LLMs to other languages in Appendix~\ref{otherlanguageLLMs}.

\subsection{Instruction Data Management}
\label{datamanagement}
As discussed above, there are extensive approaches focusing on generating high-quality instructions from different sources. 
Naturally, it becomes critical to effectively manage all of these instruction data in the LLMs alignment.

\paragraph{Instruction Implications}
Several studies focus on the implications of instruction data. ~\citet{DBLP:journals/corr/abs-2304-07854} demonstrate that an increment in the total count of training instructions can be advantageous for standard NLP tasks (e.g., information extraction, classification, Closed QA, summarization). Yet, it bears negligible influence on complex reasoning tasks such as Math, Code, CoT, and Brainstorming. Intriguingly, ~\citet{muennighoff2023scaling} discover that adding approximately 50\% of programming instructions not only leaves unaffected the general conversational performance but also enhances the reasoning prowess of LLMs. In parallel,~\citet{ghosal2023flacuna} observe that integrating FLAN-style instructions with synthetic instructions from ChatGPT/GPT-4 effectively enhances LLMs' reasoning and problem-solving capacity.

\citet{wang2023far} conduct a comprehensive analysis of the impacts of various instructions derived from different sources on factual knowledge, reasoning, coding, multilingual, and open-ended scenarios. They also reveal that instructions pertaining to CoT and Coding are vital for augmenting the reasoning capability of LLMs. Additionally, they ascertain that different instructions can affect different LLM capabilities. Therefore, a composite of all instruction types empowers the corresponding LLMs to reach their better overall performance, hinting at the need for more advanced instruction collection techniques and technologies.

\paragraph{Instruction Quantity}
Another critical question in instruction data management is the optimal quantity of instruction data required for effective LLM alignment. ~\citet{alshikh2023becoming} address this question by introducing a novel early-stopping criterion known as \textbf{IFS}. The premise of \textbf{IFS} rests on the observation that, given an input textual prefix, foundational LLMs typically predict ensuing tokens and generate "continuation-like" outputs, while fully instruction-tuned LLMs interpret the input prefix as questions, thereby generating "answer-like" outputs. \textbf{IFS} is quantified as the proportion of "answer-like" outputs within all its outputs given the instructions. The researchers train an external classifier to discriminate between "continuation-like" and "answer-like" outputs, concluding that LLaMA necessitates approximately 8K instructions to achieve a high IFS score. More instructions could potentially induce a semantic shift in the foundational LLMs.~\citet{zhou2023lima} similarly discern that merely 6K high-quality instructions suffice to align with human preferences.
Motivated by these findings, researchers are investigating high-quality instruction selection.~\citet{cao2023instruction} aim to identify predictive features of high-quality instructions. Initially, they extract representative features from the instruction dataset, then utilize these instructions to fine-tune LLMs. The feature importance is based on the model's performance. Their experiments demonstrate the better performance of LLMs trained on the resultant instructions.
Differently,~\citet{chen2023alpagasus} propose using ChatGPT to directly assess the quality of instructions by assigning scores. They report that the LLM trained on the top 9K instructions notably outperforms those trained on the complete set of 52K Alpaca instructions.

\section{Alignment Training}
\label{training}
After collecting instructions from various sources, we then consider using these data to fine-tune existing foundational LLMs to align with human. The native solution is Supervised Fine-Tuning (SFT). Specifically, given instruction input $x$, SFT calculates the cross-entropy loss over the ground-truth response $y$ as follows:
\begin{align}
    L_{ft} = -\sum_t \log P_{LLM}(y_{i',t}|x,y_{i',<t})
\end{align}
Essentially, SFT helps LLMs to understand the semantic meaning of prompts and make meaningful responses.
The main limitation of SFT is that it only teaches LLMs about the best responses and cannot provide fine-grained comparisons to sub-optimal ones. However, it is worth noting that SFT objective or SFT model parameters has also been integrated into many human preference training objective to regularize and stabilize the training process of LLMs.
We summarize the research efforts built on top of SFT into: \emph{Online human preference training}, \emph{Offline human preference training} and \emph{Parameter-effective fine-tuning solutions}.

\subsection{Online Human Preference Training}
\label{onlinetraining}
Reinforcement learning from Human Feedback (RLHF)~\cite{DBLP:conf/nips/Ouyang0JAWMZASR22} is designed to learn the human preference signals from external reward models under the PPO framework. Specifically, RLHF consists of three main stages:
\begin{itemize}
  \item \textbf{Step 1:} Collecting a high-quality instruction set and conducting SFT of pre-trained LLMs. 
  \item \textbf{Step 2:} Collecting manually ranked comparison response pairs and train a reward model ${\rm I\!R}$ to justify the quality of generated responses.
  \item \textbf{Step 3:} Optimizing the SFT model (policy) under the PPO reinforcement learning framework with reward calculated by ${\rm I\!R}$.
\end{itemize}
In Step 3, to mitigate over-optimization issues,~\citet{DBLP:conf/nips/Ouyang0JAWMZASR22} add a KL-divergence regularization between the current model weight and the SFT model weight obtained in Step 1. 
However, despite being effective in learning human preferences, PPO training is difficult in implementation and stable training.
Therefore, ~\citet{dong2023raft} try to remove the PPO training in the above process and propose 
a novel Reward rAnked FineTuning (\emph{RAFT}) method, which uses an existing reward model to select the best set of training samples based on the model outputs. Specifically, \emph{RAFT} first samples a large batch of instructions, then uses the current LLMs to respond to these instructions. These data are then ranked by the reward model and only 
top $\frac{1}{k}$ instances are applied for SFT. 
\emph{RAFT} can also be used in offline human preference learning where the global instruction set is continually updated with the top-ranked instructions in each batch. This contiguously updates the global instruction set to improve training data quality at each step.

\subsection{Offline Human Preference Training}
\label{offlinetraining}
Although the above online algorithms have been shown effective in learning human preference, implementing these algorithms could be non-trivial because its training procedure requires interaction between policy, behavior policy, reward, and value model, which requires many hyper-parameters to be tuned to achieve better stability and performance. To avoid this issue, researchers also explore learning human preferences in an offline fashion.

\subsubsection{Ranking-based Approach}
As human preferences are often expressed as a ranking result over a set of responses, some research efforts directly incorporate the ranking information into the LLMs fine-tuning stage. 
~\citet{rafailov2023direct} propose \emph{Direct Preference Optimization} (DPO), which implicitly optimizes the same objective as existing RLHF algorithms (i.e., reward function with a KL-divergence term) discussed above. Specifically, the DPO training objective can be written as:
\begin{equation}
    \mathcal{L}_\text{DPO} = \log \sigma \left[\beta \log (\frac{\pi_{\theta}(y_w\mid x)}{\pi_{\text{SFT}}(y_w\mid x)} \cdot \frac{\pi_{\text{SFT}}(y_l\mid x)}{\pi_{\theta}(y_l\mid x)})\right]
\end{equation}
where $(x, y_w, y_l)$ is one instruction and two of the corresponding outputs with $y_w$ ranked higher than $y_l$. 
Similarly,~\citet{song2023preference} propose \emph{Preference Ranking Optimization} (PRO) method, an extended version of reward model training objective proposed in~\citet{ziegler2019finetuning}, to further fine-tune LLMs to align with human preference. Given instruction $x$ and a set of responses with human preference order $y^1\succ y^2\succ \cdots \succ y^n$, the objective can be defined as follows:
\begin{equation}
\mathcal{L}_{\text{PRO}} = -\sum_{k=1}^{n-1} \log \frac{\exp\left(\pi_{\theta}(y^k\mid x)\right)}{\sum_{i=k}^{n}\exp\left(\pi_{\theta}(y^i\mid x)\right)}
\end{equation}
PRO also adds SFT training objective for the regularization purpose.
Instead of adapting the reward training objective,~\citet{zhao2023calibrating} 
take the first step to calibrate the sequence likelihood using various ranking functions, including rank loss, margin loss, list rank loss~\cite{liu-etal-2022-brio} and expected rank loss~\cite{edunov-etal-2018-classical}. In addition, they also explore to use SFT training objective and KL-divergence as the regularization term. The experiment results on various text generation tasks show that the rank loss with the KL-divergence term performs the best.
However, this paper only uses the BERTScore~\cite{Zhang*2020BERTScore:} between each candidate output and the ground-truth reference to simulate human preferences and they only conduct experiment on small pre-trained language models (i.e., no larger than 2B).~\citet{yuan2023rrhf} propose RRHF, which further optimizes LLaMA-7B to align with human preferences using a similar framework described above. RRHF is based on the list rank loss, but removes the margin terms based on the empirical results. In addition, different from~\citet{liu-etal-2022-brio}, RRHF finds that the SFT training objective is more effective and efficient than KL-divergence in preventing LLMs from over-fitting. These results show that different ranking strategies should be adapted for LLMs with different size.


\begin{figure}[!ht]
    \centering
    \includegraphics[width=0.95\columnwidth]{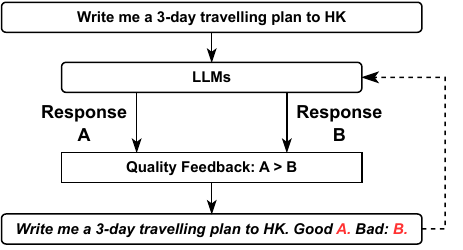}
    \caption{The overview of the Chain of Hindsigt (CoH) method. Responses with different quality are associated with different prefix. The CoH training loss is only applied on model output tokens (highlighted by \textcolor{red}{red}).}
    \label{fig:coh}
\end{figure}

\subsubsection{Language-based Approach}
As reinforcement learning algorithms are hard to optimize and LLMs have strong text understanding ability, some works propose to directly use natural language to inject human preference via SFT.~\citet{openchat} introduce the concept of ``conditional behavior cloning'' from offline reinforcement learning literature~\cite{nguyen2022conserweightive} to train LLMs to distinguish high-quality and low-quality instruction responses. Specifically, they design different language-based prefixes for different quality responses (e.g., high-quality response with ``Assistant GPT4:'' and low-quality response with ``Assistant GPT3:''). This approach can effectively leverage both low- and high-quality training data to align LLMs with humans. Chain of Hindsight (CoH)~\cite{liu2023languages}, on the other hand,  directly incorporates human preference as a pair of parallel responses discriminated as low-quality or high-quality using natural language prefixes. As shown in Figure~\ref{fig:coh}, after assigning human feedback to each model output, CoH concatenates the input instructions, LLMs outputs, and the corresponding human feedback together as the input to LLMs. Note that CoH only applies the fine-tuning loss to the actual model outputs, rather than the human feedback sequence and the instructions. During inference, CoH directly puts position feedback (e.g., good) after the input instructions to encourage the LLMs to produce high-quality outputs. It is worthnoting that, similar to~\citet{Liu2022TowardsBF,DBLP:conf/nips/Ouyang0JAWMZASR22}, CoH also incorporates SFT objectives and random words masking to prevent LLMs from over-fitting. 

Alternative approach is to 
explicitly incorporate revision-based instructions into LLMs training. Some preliminary studies have shown that many existing state-of-the-art LLMs have the capability to improve the quality of their responses when explicitly prompting them to do so~\cite{chen2023teaching}. 
Motivated by these findings,~\citet{liu2022second} recommend training LMs to produce edit operations between source (i.e., low-quality responses) and target (i.e., high-quality responses) sequences, which are subsequently integrated into a dynamic programming framework.
~\citet{liu2023training} propose a novel type of instruction called \emph{realignment}, designed to revise responses based on previously generated low-quality feedback and instructions. This compiled data is employed to instruct LLMs to self-correct when they generate bad responses.
Similarly, ~\citet{selfee2023} accumulate a multi-turn dialogue corpus utilizing this self-correction mechanism built with the ChatGPT models. Each dialogue starts with standard instructions, such as those from the Stanford Alpaca dataset. After ChatGPT has responded to the initial instructions, further revisions are explicitly requested until ChatGPT elects to terminate. They found that LLMs trained using these dialogues demonstrated an effective capacity to elevate the quality of their own responses.


\subsection{Parameter-Effective Training}
\label{parametereffective}
Directly fine-tuning all parameters in large language models (LLMs) would theoretically enable these models to adhere to provided instructions. However, this approach demands not only substantial computational resources, such as vast GPU memory but also extensive datasets for instruction training. In an effort to mitigate both computational and data requirements for constructing instruction-following LLMs, one potential route is the implementation of \emph{Parameter-Effective Fine-tuning} strategies. 
Specifically, these methods froze the major part of LLM parameters and only train a limited set of additional parameters. 

\paragraph{Supplementary Parameters}
Building upon this strategy, prefix tuning~\cite{li-liang-2021-prefix} and prompt tuning~\cite{lester-etal-2021-power} are inspired by the successful application of textual prompts in pre-trained language models~\cite{NEURIPS2020_1457c0d6}. 
These methods either prepend trainable tokens to the input layer or each hidden layer, leaving the parameters of LLMs frozen during fine-tuning. Subsequently, ~\citet{he2022towards,chen2023parameterefficient} consolidated these strategies into unified frameworks, fostering more effective solutions for parameter-efficient fine-tuning. 

\paragraph{Shadow Parameters}
While the above methodologies introduce supplementary parameters to LLMs, the following methods focus on training the weight representing model parameter variance without modifying the number of total model parameters during inference.
For instance, Low-Rank Adaptation (LoRA)~\cite{hu2022lora} suggests the addition of pairs of rank-decomposition trainable weight matrices (i.e., update matrices) to the existing weights, which are kept frozen. 
For example, given a neural layer $h=W_0x$, LoRA modifies the forward pass as follows:

\begin{equation}
    h=W_0x+BAx
\end{equation}
where $W_0 \in \mathbb{R}^{d\times k}$, $B \in \mathbb{R}^{d\times r}$, $A \in \mathbb{R}^{r\times k}$, with the rank $r \ll \min(d,k)$. LoRA only updates the parameters of $A$ and $B$ during  training.
Despite being effective, LoRA equally allocates parameter budgets over the whole LLMs, ignoring the varying importance of different weight parameters.~\citet{zhang2023adaptive} propose AdaLoRA to combat this issue. Specifically, AdaLoRA first calculates the parameter importance using the training gradient and then determines the $r$ values for different parameters matrix.~\citet{dettmers2023qlora} propose QLoRA that further improves over LoRA by reducing memory usage, enabling a 65B LLM to be fine-tuned using a single 48G GPU. Specifically, QLoRA quantizes the transformer backbone model to 4-bit precision and uses paged optimizers to handle memory spikes. 

\paragraph{Trade-offs For Parameter-efficient Training}
There are some successful applications of parameter-efficient training technologies, including the \emph{Alpaca-LoRA} project~\footnote{\url{https://github.com/tloen/alpaca-lora}}, which is based on the Hugging Face's PEFT library~\cite{peft} to train Alpaca using a single commercial GPU and~\citet{DBLP:journals/corr/abs-2304-01196}, which apply LoRA to all linear layers in LLaMA to improve its adaption capabilities. However, such an effective training approach could also result in under-fitting issues.~\citet{sun2023comparative} find that given the same set of training instructions, LLMs with LoRA perform worse than the fully fine-tuned ones. Furthermore, they also show that when using LoRA, it is preferable to use larger LLMs than larger training instruction datasets because the former solution uses less training costs and achieves better performance than the later one.

\section{Alignment Evaluation}
\label{eval}
After collecting instructions and training LLMs on these instructions, we finally consider the evaluation for alignment quality. In this section, we will discuss benchmarks used for evaluation in Section~\ref{evalbenchmark} and the evaluation protocols in Section~\ref{evalparadigm}.

\subsection{Evaluation Benchmarks}
\label{evalbenchmark}
There are various benchmarks to evaluate the aligned LLMs. In general, these benchmarks can be categorized into \emph{Closed-set Benchmarks} and \emph{Open-set Benchmarks}. The former type  focuses on evaluating the skills and knowledge of aligned LLMs, while the latter type often concentrates on the open scenarios where there are no standardized answers.

\subsubsection{Closed-set Benchmarks}
The closed-set benchmarks mostly include testing instances whose possible answers are predefined and limited to a finite set (e.g., multiple choices). We discuss some of the most commonly used benchmarks below. We refer readers to~\citet{chang2023survey} for more comprehensive introduction of LLMs' evaluation benchmarks. 

\paragraph{General Knowledge}
MMLU~\cite{hendrycks2021measuring} is an English-based benchmark to evaluate LLMs knowledge in zero-shot and few-shot settings. It comprehensively includes questions from the elementary level to an advanced professional level from 57 subjects including STEM, the humanities, the social sciences, etc. The granularity and breadth of the subjects make MMLU ideal for identifying LLMs’ blind spots.
There are also several benchmarks attempting in evaluating the general knowledge in Chinese LLMs.
C-MMLU~\cite{li2023cmmlu}, C-Eval~\cite{huang2023ceval}, M3KE~\cite{liu2023m3ke} and AGIEval~\cite{zhong2023agieval} 
are all Chinese counterparts of MMLU that include diverse sets of questions from multiple subjects with different difficulty levels from various Chinese standardized exams, including Chinese college entrance exams, advanced maths competitions and law exams.
The KoLA benchmark~\cite{yu2023kola} is proposed to evaluate the general real-world knowledge of LLMs.

\paragraph{Reasoning}
Reasoning is a fundamental type of human intelligence that are crucial in solving complicated tasks. Interestingly, research find that LLMs have exhibit emergent behaviors, including the reasoning ability, when they are sufficiently large. Thus, there are several benchmarks in evaluating the ability of arithmetic, commonsense, and symbolic reasoning for LLMs. GSM8K~\cite{cobbe2021training} and Maths~\cite{hendrycks2021measuring} are designed to evaluate the arithmetic reasoning ability for LLMs. CSQA~\cite{talmor-etal-2019-commonsenseqa} and StrategyQA~\cite{geva-etal-2021-aristotle} are proposed to evaluate the commonsense reasoning ability which requires the LLMs to use daily life commonsense to infer in novel situations.~\citet{wei2022chain} propose two novel tasks, Last Letter Concatenation and Coin Flip and measure the Symbolic reasoning ability that involves the manipulation of symbols according to formal rules. BBH~\cite{suzgun2022challenging}, a challenging subset of BIG-Bench~\cite{srivastava2023beyond}, focus on evaluating a wide range of reasoning skills, such as Date Understanding, Word Sorting, and Causal Judgement. 

\paragraph{Coding}
HumanEval~\cite{chen2021evaluating}, HumanEval+~\cite{liu2023your}, and MBPP~\cite{austin2021program} are extensively used benchmarks to evaluate the coding skills of LLMs. They encompass a vast collection of Python programming problems and corresponding test cases to automatically verify the code generated by Code LLMs. The DS-1000 benchmark~\cite{Lai2022DS1000} comprises 1,000 distinct data science workflows spanning seven
libraries. It assesses the performance of code generations against test cases and supports two evaluation modes: completion and insertion.

\subsubsection{Open-ended Benchmarks}
In contrast to the closed-set benchmarks, the responses to open-set benchmarks can be more flexible and diverse, where aligned LLMs are usually given chatting questions or topics that do not have any fixed reference answers. Early attempts of open-ended benchmarks, such as Vicuna-80~\cite{vicuna2023}, Open-Assistant-953~\cite{Kopf2023OpenAssistantC}, User-Instructions-252~\cite{DBLP:journals/corr/abs-2212-10560}, often leverage a small number of syntactic instructions from LLMs as testing instances. All evaluation candidate LLMs are prompted with the same instructions to provide responses, which are then evaluated against human-based or LLMs-based evaluators. However, these types of benchmarks can only provide comparison several LLMs at a time, making it challenging to reveal a fair comparison among a board range of LLMs, as well as incremental updates when new LLMs become available. AlpacaEval~\cite{dubois2023alpacafarm} tackles this issue by reporting the \emph{Win Rate} of the LLMs candidate to 
the reference LLM \emph{text-davinci-003}. Accordingly, LLMs with higher \emph{Win Rate} are generally better than the ones with lower \emph{Win Rate}. MT-Bench~\cite{zheng2023judging} further increases the evaluation difficulty by proposing 80 multi-turn evaluation instances and wishes LLMs could effectively capture context information in previous turns. FLASK~\cite{Ye2023FLASKFL} proposed to provide fine-grained evaluation towards aligned LLMs. FLASK includes  1,700 instances from 120 datasets. Each testing instance is labelled with a set of 12 foundational and essential ``alignment skills'' (e.g., logical thinking, user alignment, etc.). Accordingly, it is straightforward to evaluate LLMs' capabilities on these skills separately.

\subsection{Evaluation Paradigm}
\label{evalparadigm}
As open-ended benchmarks often do not have reference answers, it is essential to rely on external human or LLMs evaluators. In this section, we will introduce both human- and LLMs-based evaluation paradigm.

\subsubsection{Human-based Evaluation}

Automatic metrics, such as BLUE~\cite{papineni-etal-2002-bleu} and ROUGE~\cite{lin-2004-rouge}, require ground-truth references and have relatively low correlation with human judgments. Thus, they are not feasible for evaluating responses to open-ended questions.
To bridge this gap, human annotators are used to evaluate the quality of open-ended model responses.
~\citet{DBLP:journals/corr/abs-2212-10560,DBLP:journals/corr/abs-2304-1440} propose to evaluate the response quality in an ordinal classification setting where human annotators are instructed to categorize each response into one of the four levels (i.e., acceptable, minor errors, major errors and unacceptable), separately. However, some other research have found that such classification annotation strategy heavily depend on the subjectivity of annotators, which can result in poor inter-rater reliability~\cite{KALPATHYCRAMER20162345}. Accordingly 
~\citet{alpaca} propose to use a pairwise comparison framework for evaluating the output quality of two LLMs systems. Given the instruction inputs and two model outputs, the human annotators are asked to select a better one. Furthermore, to accurately evaluate multiple LLMs,~\citet{zheng2023judging,dettmers2023qlora} further introduce  the Elo rating system which calculates the relative skill levels of players in zero-sum games such as chess games. Specifically, in Elo system, the player scores are updated based on the result of each pairwise comparison and the current player scores.

\subsubsection{LLMs-based Evaluation}
While human evaluations are often of high quality, it could be inefficient and expensive. In addition, the increasing quality of generated text from LLMs makes it more challenging for human annotators to distinguish between human-written and LLM-generated text in the open-ended NLP tasks~\cite{Clark2021AllT}. Given the strong text capability of LLMs, recent studies propose to incorporate LLMs into the output text evaluation in various NLP tasks without additional expensive references and human efforts.~\citet{tang2023not} propose to improve the traditional automatic metrics by increasing the number of references via LLMs-based paraphrasing systems. However, such method still requires one reference for each evaluation instance. In contrast, ~\citet{liu2023gpteval,fu2023gptscore,chen2023exploring,chiang2023can} propose to 
directly use LLMs to evaluate the generated text quality without a single reference in a wide range of Natural Language Generation (NLG) tasks. Specifically, they construct complicated input instructions with tasks background and evaluation rules and prompt LLMs to follow these evaluation instructions to provide scores for output text. There are also some research efforts that propose LLMs-based evaluation framework for specific NLG tasks, including text summarization~\citet{gao2023human}, code generation~\cite{zhuo2023large}, open-ended QA~\cite{bai2023benchmarking} and conversations~\cite{lin2023llm}. Due to the flexibility of prompts, it is also possible to conduct multi-dimensional evaluation towards the generated text~\cite{lin2023llm,fu2023gptscore}.~\citet{min2023factscore,zha2023alignscore} propose to evaluate factual correctness using both closed-sourced and open-sourced LLMs. Similar to human evaluation, there are also research efforts in explicitly prompting LLMs to conduct pairwise comparisons. To compare the capabilities of two LLMs, instead of assigning scores separately,~\citet{dubois2023alpacafarm,zheng2023judging} explicitly to prompt GPT-4 to select the better response for the same instruction inputs.

\paragraph{LLMs Evaluation Bias}
Despite LLMs achieve impressive consistency with human judgment, 
~\citet{wang2023large} find that such LLM-based evaluation paradigm suffers from a positional bias and those strong LLMs (i.e., GPT-4) tend to assign higher scores to the first appeared candidates. To calibrate such bias, they propose 
to \textbf{a)} repeat the LLM evaluation process multiple times with different candidate ordering and \textbf{b)} explicitly prompt LLMs to provide chain-of-thoughts for the evaluation before assigning the actual score.~\cite{Wu2023StyleOS} find that LLM-based evaluation prefer candidates with factual errors over shorter candidates and candidates with grammatical errors, despite the former one could impose greater danger than the latter ones. To address this bias, they propose a multi-dimensional Elo rating system which separately evaluates the candidates from the perspective of accuracy, helpfulness and language. Such approach allows a more comprehensive understanding towards the candidates quality than previous one-shot evaluation. Concretely,~\cite{zheng2023judging} systematically show the bias LLMs-based evaluation systems. On top of positional and length bias, they also discover Self-enhancement bias which means LLMs favor their own responses than the ones from other sources. To tackle these biases, their solutions include swapping responses, adding few-shot examples and leveraging CoT and references information. 

\paragraph{Evaluation-Specific LLM}
Despite achieving high-quality automatic evaluation results, the above approaches heavily rely on state-of-the-art closed-source LLMs (e.g., GPT-4) which could result in data privacy issues.~\cite{zheng2023judging} propose to train evaluation-specific LLMs. PandaLM~\cite{wang2023pandalm} is such a specialized evaluation LLMs by fine-tuning LLaMA-7B using around 300K high-quality synthetic evaluation instructions generated from GPT-3.5. Specifically, they first collect large volumes of instructions as well as outputs from a diverse range of open-sourced LLMs, such as LLaMA-7B and Bloom-7B. They then prompt GPT-3.5 to analysis and evaluate the quality of a pair of outputs. Their results on human-annotated meta-evaluation shows that, despite bebing much smaller, PandaLM achieves on-par evaluation performance comparing to GPT-3.5 and GPT-4.

\section{Challenges and Future Directions}
\label{challengesdirection}
The development of LLM alignment is still in a rudimentary stage and thus leaves much room for improvement. In this section, we summarize existing important research efforts of aligning LLMs with human in Table~\ref{llmsummary}. Below, we will discuss some of the challenges as well as the corresponding future research directions.

\begin{table*}[!ht]
    \centering
    \resizebox{\linewidth}{!}{
    \begin{tabular}{c|cccccccccc} \toprule
Aligned LLM & Size & Lang. & Initial LLMs & Training & Self Instruction & NLP Benchmarks & Human Annotations & Human Eval & Auto. Benchmark Eval         & LLM Eval                  \\ \midrule
Alpaca~\cite{alpaca} & 7B & EN & LLaMA & SFT & Text-Davinci-003 & \xmark & \xmark & Author Verification & \xmark & \xmark \\
Vicuna~\cite{vicuna2023} & 7B, 13B, 33B & EN & LLaMA & SFT & GPT-3.5 & \xmark & 70K ShareGPT & \xmark & \xmark & \emph{Vicuna-80} \\
GPT4ALL~\cite{gpt4all}     & 6B, 13B & EN & \makecell{LLaMA \\ GPT-J} & SFT & \xmark                & Bloomz-P3      & \makecell{OIG, ShareGPT, Dolly \\ Stack Overflow}                & \xmark     & Common Sense Reasoning & \xmark \\
LLaMA-GPT4~\cite{peng2023instruction} & 7B & EN, CN & LLaMA & SFT & \makecell{Text-Davinci-003 \\ GPT-4} & \xmark & \xmark & \makecell{\emph{User-Instructions-252} \\ Pairwise, AMT} &  Unnatural Instructions & \emph{Vicuna-80} \\
Phoenix~\cite{DBLP:journals/corr/abs-2304-10453} & 7B, 13B & Multilingual & \makecell{LLaMA \\ BLOOMZ} & SFT & \makecell{GPT-3.5 Multilingual \\ and Dialogue Data} & \xmark & ShareGPT & Volunteers & \xmark & GPT-3.5, GPT-4 \\
UltraLLaMA~\cite{ding2023enhancing} & 13B & EN &  LLaMA & SFT & GPT-3.5 Dialogue Data & \xmark & \xmark & \xmark & Truthful QA & \makecell{GPT 3.5 \emph{Vicuna-80} \\ 300 diverse questions} \\
Baize~\cite{DBLP:journals/corr/abs-2304-01196} & 7B, 13B, 30B & EN & LLaMA & Revision, LoRA & GPT-3.5 self-Chat Data & \xmark & Quora Questions & \xmark & \xmark & GPT-4 \\
WizardLM~\cite{xu2023wizardlm} & 7B, 13B, 30B & EN & LLaMA & SFT & \makecell{GPT-3.5, Alpaca \\ Complex Instructions} & \xmark & ShareGPT & \makecell{10 Annotators \\ Pairwise Comparison} & \xmark & GPT-4, \emph{WizedLM-218} \\
WizardCoder~\cite{luo2023wizardcoder} & 15B & EN, Code & StarCoder & SFT & \makecell{GPT-3.5, Code Alpaca \\ Complex Instructions} & \xmark & \xmark & \xmark & \makecell{HumanEval, MBPP \\ HumanEval+, DS-1000} & \xmark \\
OpenChat~\cite{openchat} & 13B & EN & LLaMA & Language & \xmark & \xmark & \makecell{GPT 3.5 \& GPT4 \\ ShareGPT} & \xmark & MMLU & GPT-4 \\
Guanaco~\cite{dettmers2023qlora} & 13B, 33B, 65B & EN & LLaMA & QLoRA & \makecell{Alpaca, SELF-INSTRUCT \\ Unnatural instructions} & FLAN & Chip2 & Elo, \emph{Vicuna-80} & MMLU & \makecell{Elo, \emph{Vicuna-80} \\ \emph{Open-Assistant-953}} \\
MPT-chat~\cite{MosaicML2023Introducing} & 13B, 30B & EN & MPT & SFT & \makecell{GPTeacher, Guanaco \\ Baize Instructions} & \xmark & Vicuna ShareGPT & \xmark & MMLU & GPT4, MT-bench \\
FLACUNA~\cite{ghosal2023flacuna} & 13B & EN & Vicuna & LoRA & Alpaca, Code Alpaca & FLAN & ShareGPT & \xmark & \makecell{MMLU, BBH, DROP \\ CRASS, HumanEval} & GPT 3.5, IMPACT \\
Bactrian-X~\cite{bactrian} & 7B & Multilingual & \makecell{LLaMA \\ BLOOMZ} & LoRA & \makecell{Alpaca \\ Google Translation} & \xmark & \xmark & \xmark & \makecell{XCOPA, XStoryCloze \\ XWinograd, SentimentX} & \makecell{GPT 4 \\ Multilingual \emph{Vicuna-80}}  \\
Ocra~\cite{mukherjee2023orca} & 13B & EN & LLaMA & SFT & \xmark & FLAN & \xmark & \xmark & AGIEval, BBH & \makecell{GPT-4, \emph{Vicuna-80} \\ \emph{WizedLM-218}, \emph{Awesome-164}} \\
Phi-1~\cite{gunasekar2023textbooks} & 350M, 1.3B & EN, Code & Phi-1-base & SFT & \makecell{GPT-3.5 \\ Synthetic Textbook} & \xmark & \makecell{Python, The Stack \\ Stack Overflow} & \xmark & HumanEval & GPT-4 Grading \\
Chinese Alpaca~\cite{cui2023efficient} & 7B, 13B, 33B & EN, CN & Chinese LLaMA & LoRA & \makecell{STEM \\ Org. and Trans.  Alpaca} & pCLUE & \xmark & \xmark & C-Eval & \xmark \\
Lion~\cite{Jiang2023LionAD} & 7B, 13B & EN & LLaMA & SFT & \makecell{Alpaca \\ GPT 3.5 Adv. Instruction} & \xmark & \xmark & \makecell{HHH \\ \emph{User-Instructions-252}} & \xmark & GPT-4, \emph{Vicuna-80} \\
Stable Alignment~\cite{liu2023training} & 7B & EN & Alpaca & SFT & \makecell{GPT-3.5 \\ Social Aligned Instructions} & \xmark & \xmark & \xmark & \xmark & \makecell{GPT-4 \\ HHH, HHH-A} \\
Dromedary~\cite{Sun2023PrincipleDrivenSO} & 65B & EN & LLaMA & SFT & LLaMA-65B, Self-Align & \xmark & \makecell{175 Munnal Examples \\ 16 Principle Rules} & \xmark & TruthfulQA, BBH & GPT-4, \emph{Vicuna-80} \\
Dolly-v2~\cite{DatabricksBlog2023DollyV2} & 3B, 7B, 12B & EN & Pythia & SFT & \xmark & \xmark & \emph{databricks-dolly-15k} & \xmark & \emph{LLM Harness} & \xmark \\
Selfee~\cite{selfee2023} & 7B, 13B & EN & LLaMA & Revision & \makecell{GPT 3.5 Self-Improve \\ Alpaca} & FLAN, Maths, Code & ShareGPT & \xmark & \xmark & GPT-4, \emph{Vicuna-80} \\
\textsc{T\"ulu}~\cite{wang2023far} & 7B, 13B, 30B, 65B & EN & LLaMA & SFT & \makecell{Alpaca, Code Alpaca \\ GPT4-Alpaca, Self-instruct} & FLAN, CoT & \makecell{Dolly, ShareGPT \\ Open Assistant} & \makecell{Acceptability \\ Pairwise Comparison} & \makecell{MMLU, GSM, BBH \\ TydiQA, Codex-Eval} & \makecell{GPT4 on \emph{Vicuna-80}, Koala \\ Open Assistant Benchmarks}\\
Koala~\cite{koala_blogpost_2023} & 13B & EN & LLaMA & Language & Alpaca & \xmark & \makecell{OIG, HC3, Anthropic HH \\ OpenAI WebGPT, Summary} & \makecell{100 AMT Annotators \\ on \emph{Alpaca and Koala Test}} & \xmark & \xmark \\
Bayling~\cite{Zhang2023BayLingBC} & 7B, 13B & Multilingual & LLaMA & SFT & \makecell{Alpaca \\ GPT 3.5  Interactive Translation} & \xmark & ShareGPT & Translation Quality & \makecell{WMT22 Multilingual Translation \\  Lexically Constrained Translation} & \xmark \\
Wombat~\cite{yuan2023rrhf} & 7B & EN & Alpaca & Rank & \makecell{Alpaca \\ ChatGPT Ratings} & \xmark & \emph{Helpful and Harmless} & \xmark & \xmark & GPT-4, \emph{Vicuna-80} \\
Lamini-lm~\cite{DBLP:journals/corr/abs-2304-1440} & 0.7B & EN & T5-Flan & SFT & \makecell{Alpaca \\ Self-instruct} & P3, FLAN & \xmark & Human Rating & \emph{LLM harness} & \xmark \\
\bottomrule
\end{tabular}}
\caption{An overview of popular aligned LLMs, including their Size, supported languages, initial LLMs, alignment training method, alignment data, and alignment evaluation.}
\label{llmsummary}
\end{table*}

\paragraph{Fine-grained Instruction Data Management}
While research on LLMs alignment have been unprecedentedly active, many of these research efforts propose to leverage training instructions from diverse sources, making it challenging to fairly compare among different methods. As discussed in Section~\ref{datamanagement}, there are some interesting findings about the implication of particular instruction dataset. For example, FLAN and programming instructions can improve reasoning capability aligned LLMs~\cite{ghosal2023flacuna} and ShareGPT general performs well across a wide range of benchmarks~\cite{wang2023far}. 
However, there are still many issues in other aspects of instruction data management remaining unclear, including the optimal quality control towards instruction data, optimal instruction training sequence, how to effectively mix-up different instructions. These research efforts could finally enable fine-grained instruction management, allowing researchers and practitioners to construct high-quality instruction data.


\paragraph{LLMs Alignment for non-English Languages} 
Most of existing research in LLMs alignment are English-dominated. While many approaches, such as complex instruction generation~\cite{xu2023wizardlm} and explanation tuning~\cite{mukherjee2023orca}, are language-agnostic, they only explore English-based prompts and it is unclear how well these prompts perform when adapting to other languages, severely hindering the application of LLMs to non-English regions. It is interesting to see \emph{1)} how these alignment technologies perform in various languages, in particular low-resource languages, and \emph{2)} how to effectively transfer the effect of LLMs alignment across different languages.

\paragraph{LLMs Alignment Training Technologies}
As shown in Table~\ref{llmsummary}, most of existing 
aligned LLMs are based on the simple SFT technology. However, SFT does not explicitly incorporate human preference into LLMs. As a result, aligning LLMs solely based on SFT could require a lot more instruction data and training resources. In general, there is a lacking of comprehensive investigation over the effect of various training technologies to incorporate human preference into LLMs. 
Thus, it is critical to come up with resource-constrained LLM alignment training framework where certain alignment resources are given at a certain level (e.g., maximum 10K instructions, 5 hours training time, etc.), allowing researchers and practitioners to verify the effectiveness of various training methods. As increasing number of instruction data have become available, this exploration could further 
promote effective and environmental-friendly LLMs alignment solutions.

\paragraph{Human-in-the-loop LLMs Alignment Data Generation}
Table~\ref{llmsummary} has shown that ShareGPT data has been widely adapted for LLMs alignment. The preliminary analysis in~\citet{wang2023far} also reveal that   ShareGPT  performs consistly well across a wide range of NLP tasks. These results indicate that human is still a key factor in improving LLMs alignment quality. Different from traditional human annotation framework where human provides annotation based on the instructions, ShareGPT is a human-in-the-loop alignment solution where human can freely determine what LLMs should generate. This shows the great potential of human-in-the-loop data generation solution in LLMs alignment. It will be interesting to explore other types of human-in-the-loop solutions to further facilitate LLMs alignment.

\paragraph{Human-LLM Joint Evaluation Framework}
Existing LLM evaluation frameworks either use LLMs for effective evaluation or leverage crowd-sourcing for high-quality evaluation. As shown in~\cite{Wu2023StyleOS,liu2023gpteval}, state-of-the-art LLMs have demonstrated similar or superior evaluation capability in various NLP tasks.
It is feasible to use LLMs as special evaluation annotators and develop LLM-human joint evaluation framework where LLMs and human are assigned with different evaluation tasks based on their own strengths to maintain both efficiency and quality of the evaluation procedure for LLM alignment .

\section{Conclusion}
\label{conclusion}
This survey provides an up-to-date review to recent advances of LLMs alignment technologies. We summarize these research efforts into \emph{Alignment Instruction Collection}, \emph{Alignment Training} and \emph{Alignment Evaluation}.  Finally, we pointed out several promising future directions for LLMs alignment. We hope this survey could provide insightful perspectives and inspire further research in improving LLMs alignment.

\bibliography{anthology,custom}

\begin{thebibliography}{109}
\expandafter\ifx\csname natexlab\endcsname\relax\def\natexlab#1{#1}\fi

\bibitem[{AlShikh et~al.(2023)AlShikh, Daaboul, Goddard, Imel, Kamble,
  Kulkarni, and Russak}]{alshikh2023becoming}
Waseem AlShikh, Manhal Daaboul, Kirk Goddard, Brock Imel, Kiran Kamble,
  Parikshith Kulkarni, and Melisa Russak. 2023.
\newblock Becoming self-instruct: introducing early stopping criteria for
  minimal instruct tuning.
\newblock \emph{arXiv preprint arXiv:2307.03692}.

\bibitem[{Anand et~al.(2023)Anand, Nussbaum, Duderstadt, Schmidt, and
  Mulyar}]{gpt4all}
Yuvanesh Anand, Zach Nussbaum, Brandon Duderstadt, Benjamin Schmidt, and Andriy
  Mulyar. 2023.
\newblock Gpt4all: Training an assistant-style chatbot with large scale data
  distillation from gpt-3.5-turbo.
\newblock \url{https://github.com/nomic-ai/gpt4all}.

\bibitem[{Austin et~al.(2021)Austin, Odena, Nye, Bosma, Michalewski, Dohan,
  Jiang, Cai, Terry, Le et~al.}]{austin2021program}
Jacob Austin, Augustus Odena, Maxwell Nye, Maarten Bosma, Henryk Michalewski,
  David Dohan, Ellen Jiang, Carrie Cai, Michael Terry, Quoc Le, et~al. 2021.
\newblock Program synthesis with large language models.
\newblock \emph{arXiv preprint arXiv:2108.07732}.

\bibitem[{Bach et~al.(2022)Bach, Sanh, Yong, Webson, Raffel, Nayak, Sharma,
  Kim, Bari, Fevry, Alyafeai, Dey, Santilli, Sun, Ben-david, Xu, Chhablani,
  Wang, Fries, Al-shaibani, Sharma, Thakker, Almubarak, Tang, Radev, Jiang, and
  Rush}]{bach-etal-2022-promptsource}
Stephen Bach, Victor Sanh, Zheng~Xin Yong, Albert Webson, Colin Raffel,
  Nihal~V. Nayak, Abheesht Sharma, Taewoon Kim, M~Saiful Bari, Thibault Fevry,
  Zaid Alyafeai, Manan Dey, Andrea Santilli, Zhiqing Sun, Srulik Ben-david,
  Canwen Xu, Gunjan Chhablani, Han Wang, Jason Fries, Maged Al-shaibani, Shanya
  Sharma, Urmish Thakker, Khalid Almubarak, Xiangru Tang, Dragomir Radev, Mike
  Tian-jian Jiang, and Alexander Rush. 2022.
\newblock \href {https://doi.org/10.18653/v1/2022.acl-demo.9}
  {{P}rompt{S}ource: An integrated development environment and repository for
  natural language prompts}.
\newblock In \emph{Proceedings of the 60th Annual Meeting of the Association
  for Computational Linguistics: System Demonstrations}, pages 93--104, Dublin,
  Ireland. Association for Computational Linguistics.

\bibitem[{Bai et~al.(2023)Bai, Ying, Cao, Lv, He, Wang, Yu, Zeng, Xiao, Lyu
  et~al.}]{bai2023benchmarking}
Yushi Bai, Jiahao Ying, Yixin Cao, Xin Lv, Yuze He, Xiaozhi Wang, Jifan Yu,
  Kaisheng Zeng, Yijia Xiao, Haozhe Lyu, et~al. 2023.
\newblock Benchmarking foundation models with language-model-as-an-examiner.
\newblock \emph{arXiv preprint arXiv:2306.04181}.

\bibitem[{bench authors(2023)}]{srivastava2023beyond}
BIG bench authors. 2023.
\newblock \href {https://openreview.net/forum?id=uyTL5Bvosj} {Beyond the
  imitation game: Quantifying and extrapolating the capabilities of language
  models}.
\newblock \emph{Transactions on Machine Learning Research}.

\bibitem[{Brown et~al.(2020)Brown, Mann, Ryder, Subbiah, Kaplan, Dhariwal,
  Neelakantan, Shyam, Sastry, Askell, Agarwal, Herbert-Voss, Krueger, Henighan,
  Child, Ramesh, Ziegler, Wu, Winter, Hesse, Chen, Sigler, Litwin, Gray, Chess,
  Clark, Berner, McCandlish, Radford, Sutskever, and
  Amodei}]{NEURIPS2020_1457c0d6}
Tom Brown, Benjamin Mann, Nick Ryder, Melanie Subbiah, Jared~D Kaplan, Prafulla
  Dhariwal, Arvind Neelakantan, Pranav Shyam, Girish Sastry, Amanda Askell,
  Sandhini Agarwal, Ariel Herbert-Voss, Gretchen Krueger, Tom Henighan, Rewon
  Child, Aditya Ramesh, Daniel Ziegler, Jeffrey Wu, Clemens Winter, Chris
  Hesse, Mark Chen, Eric Sigler, Mateusz Litwin, Scott Gray, Benjamin Chess,
  Jack Clark, Christopher Berner, Sam McCandlish, Alec Radford, Ilya Sutskever,
  and Dario Amodei. 2020.
\newblock \href
  {https://proceedings.neurips.cc/paper_files/paper/2020/file/1457c0d6bfcb4967418bfb8ac142f64a-Paper.pdf}
  {Language models are few-shot learners}.
\newblock In \emph{Advances in Neural Information Processing Systems},
  volume~33, pages 1877--1901. Curran Associates, Inc.

\bibitem[{Cao et~al.(2023)Cao, Kang, and Sun}]{cao2023instruction}
Yihan Cao, Yanbin Kang, and Lichao Sun. 2023.
\newblock Instruction mining: High-quality instruction data selection for large
  language models.
\newblock \emph{arXiv preprint arXiv:2307.06290}.

\bibitem[{Chang et~al.(2023)Chang, Wang, Wang, Wu, Zhu, Chen, Yang, Yi, Wang,
  Wang et~al.}]{chang2023survey}
Yupeng Chang, Xu~Wang, Jindong Wang, Yuan Wu, Kaijie Zhu, Hao Chen, Linyi Yang,
  Xiaoyuan Yi, Cunxiang Wang, Yidong Wang, et~al. 2023.
\newblock A survey on evaluation of large language models.
\newblock \emph{arXiv preprint arXiv:2307.03109}.

\bibitem[{Chen et~al.(2023{\natexlab{a}})Chen, Zhang, Shi, Li, Smola, and
  Yang}]{chen2023parameterefficient}
Jiaao Chen, Aston Zhang, Xingjian Shi, Mu~Li, Alex Smola, and Diyi Yang.
  2023{\natexlab{a}}.
\newblock \href {https://openreview.net/forum?id=XSRSWxyJIC}
  {Parameter-efficient fine-tuning design spaces}.
\newblock In \emph{The Eleventh International Conference on Learning
  Representations}.

\bibitem[{Chen et~al.(2023{\natexlab{b}})Chen, Li, Yan, Wang, Gunaratna, Yadav,
  Tang, Srinivasan, Zhou, Huang et~al.}]{chen2023alpagasus}
Lichang Chen, Shiyang Li, Jun Yan, Hai Wang, Kalpa Gunaratna, Vikas Yadav,
  Zheng Tang, Vijay Srinivasan, Tianyi Zhou, Heng Huang, et~al.
  2023{\natexlab{b}}.
\newblock Alpagasus: Training a better alpaca with fewer data.
\newblock \emph{arXiv preprint arXiv:2307.08701}.

\bibitem[{Chen et~al.(2021)Chen, Tworek, Jun, Yuan, Pinto, Kaplan, Edwards,
  Burda, Joseph, Brockman et~al.}]{chen2021evaluating}
Mark Chen, Jerry Tworek, Heewoo Jun, Qiming Yuan, Henrique Ponde de~Oliveira
  Pinto, Jared Kaplan, Harri Edwards, Yuri Burda, Nicholas Joseph, Greg
  Brockman, et~al. 2021.
\newblock Evaluating large language models trained on code.
\newblock \emph{arXiv preprint arXiv:2107.03374}.

\bibitem[{Chen et~al.(2023{\natexlab{c}})Chen, Lin, Sch{\"a}rli, and
  Zhou}]{chen2023teaching}
Xinyun Chen, Maxwell Lin, Nathanael Sch{\"a}rli, and Denny Zhou.
  2023{\natexlab{c}}.
\newblock Teaching large language models to self-debug.
\newblock \emph{arXiv preprint arXiv:2304.05128}.

\bibitem[{Chen et~al.(2023{\natexlab{d}})Chen, Wang, Jiang, Shi, and
  Xu}]{chen2023exploring}
Yi~Chen, Rui Wang, Haiyun Jiang, Shuming Shi, and Ruifeng Xu.
  2023{\natexlab{d}}.
\newblock Exploring the use of large language models for reference-free text
  quality evaluation: A preliminary empirical study.
\newblock \emph{arXiv preprint arXiv:2304.00723}.

\bibitem[{Chen et~al.(2023{\natexlab{e}})Chen, Jiang, Chen, Wang, Yu, Chen,
  Zhang, Liang, Zhang, Zhang, Li, Wan, Wang, and
  Li}]{DBLP:journals/corr/abs-2304-10453}
Zhihong Chen, Feng Jiang, Junying Chen, Tiannan Wang, Fei Yu, Guiming Chen,
  Hongbo Zhang, Juhao Liang, Chen Zhang, Zhiyi Zhang, Jianquan Li, Xiang Wan,
  Benyou Wang, and Haizhou Li. 2023{\natexlab{e}}.
\newblock \href {https://doi.org/10.48550/arXiv.2304.10453} {Phoenix:
  Democratizing chatgpt across languages}.
\newblock \emph{CoRR}, abs/2304.10453.

\bibitem[{Chiang and Lee(2023)}]{chiang2023can}
Cheng-Han Chiang and Hung-yi Lee. 2023.
\newblock Can large language models be an alternative to human evaluations?
\newblock \emph{arXiv preprint arXiv:2305.01937}.

\bibitem[{Chiang et~al.(2023)Chiang, Li, Lin, Sheng, Wu, Zhang, Zheng, Zhuang,
  Zhuang, Gonzalez, Stoica, and Xing}]{vicuna2023}
Wei-Lin Chiang, Zhuohan Li, Zi~Lin, Ying Sheng, Zhanghao Wu, Hao Zhang, Lianmin
  Zheng, Siyuan Zhuang, Yonghao Zhuang, Joseph~E. Gonzalez, Ion Stoica, and
  Eric~P. Xing. 2023.
\newblock \href {https://vicuna.lmsys.org} {Vicuna: An open-source chatbot
  impressing gpt-4 with 90\%* chatgpt quality}.

\bibitem[{Clark et~al.(2021)Clark, August, Serrano, Haduong, Gururangan, and
  Smith}]{Clark2021AllT}
Elizabeth Clark, Tal August, Sofia Serrano, Nikita Haduong, Suchin Gururangan,
  and Noah~A. Smith. 2021.
\newblock All that’s ‘human’ is not gold: Evaluating human evaluation of
  generated text.
\newblock In \emph{Annual Meeting of the Association for Computational
  Linguistics}.

\bibitem[{Cobbe et~al.(2021)Cobbe, Kosaraju, Bavarian, Chen, Jun, Kaiser,
  Plappert, Tworek, Hilton, Nakano et~al.}]{cobbe2021training}
Karl Cobbe, Vineet Kosaraju, Mohammad Bavarian, Mark Chen, Heewoo Jun, Lukasz
  Kaiser, Matthias Plappert, Jerry Tworek, Jacob Hilton, Reiichiro Nakano,
  et~al. 2021.
\newblock Training verifiers to solve math word problems.
\newblock \emph{arXiv preprint arXiv:2110.14168}.

\bibitem[{Conover et~al.(2023)Conover, Hayes, Mathur, Xie, Wan, Shah, Ghodsi,
  Wendell, Zaharia, and Xin}]{DatabricksBlog2023DollyV2}
Mike Conover, Matt Hayes, Ankit Mathur, Jianwei Xie, Jun Wan, Sam Shah, Ali
  Ghodsi, Patrick Wendell, Matei Zaharia, and Reynold Xin. 2023.
\newblock \href
  {https://www.databricks.com/blog/2023/04/12/dolly-first-open-commercially-viable-instruction-tuned-llm}
  {Free dolly: Introducing the world's first truly open instruction-tuned llm}.

\bibitem[{Cui et~al.(2023{\natexlab{a}})Cui, Yang, and
  Yao}]{chinese-llama-alpaca}
Yiming Cui, Ziqing Yang, and Xin Yao. 2023{\natexlab{a}}.
\newblock \href {https://arxiv.org/abs/2304.08177} {Efficient and effective
  text encoding for chinese llama and alpaca}.
\newblock \emph{arXiv preprint arXiv:2304.08177}.

\bibitem[{Cui et~al.(2023{\natexlab{b}})Cui, Yang, and Yao}]{cui2023efficient}
Yiming Cui, Ziqing Yang, and Xin Yao. 2023{\natexlab{b}}.
\newblock Efficient and effective text encoding for chinese llama and alpaca.
\newblock \emph{arXiv preprint arXiv:2304.08177}.

\bibitem[{Dettmers et~al.(2023)Dettmers, Pagnoni, Holtzman, and
  Zettlemoyer}]{dettmers2023qlora}
Tim Dettmers, Artidoro Pagnoni, Ari Holtzman, and Luke Zettlemoyer. 2023.
\newblock Qlora: Efficient finetuning of quantized llms.
\newblock \emph{arXiv preprint arXiv:2305.14314}.

\bibitem[{Ding et~al.(2023)Ding, Chen, Xu, Qin, Zheng, Hu, Liu, Sun, and
  Zhou}]{ding2023enhancing}
Ning Ding, Yulin Chen, Bokai Xu, Yujia Qin, Zhi Zheng, Shengding Hu, Zhiyuan
  Liu, Maosong Sun, and Bowen Zhou. 2023.
\newblock Enhancing chat language models by scaling high-quality instructional
  conversations.
\newblock \emph{arXiv preprint arXiv:2305.14233}.

\bibitem[{Dong et~al.(2023)Dong, Xiong, Goyal, Pan, Diao, Zhang, Shum, and
  Zhang}]{dong2023raft}
Hanze Dong, Wei Xiong, Deepanshu Goyal, Rui Pan, Shizhe Diao, Jipeng Zhang,
  Kashun Shum, and Tong Zhang. 2023.
\newblock Raft: Reward ranked finetuning for generative foundation model
  alignment.
\newblock \emph{arXiv preprint arXiv:2304.06767}.

\bibitem[{Dubois et~al.(2023)Dubois, Li, Taori, Zhang, Gulrajani, Ba, Guestrin,
  Liang, and Hashimoto}]{dubois2023alpacafarm}
Yann Dubois, Xuechen Li, Rohan Taori, Tianyi Zhang, Ishaan Gulrajani, Jimmy Ba,
  Carlos Guestrin, Percy Liang, and Tatsunori~B Hashimoto. 2023.
\newblock Alpacafarm: A simulation framework for methods that learn from human
  feedback.
\newblock \emph{arXiv preprint arXiv:2305.14387}.

\bibitem[{Edunov et~al.(2018)Edunov, Ott, Auli, Grangier, and
  Ranzato}]{edunov-etal-2018-classical}
Sergey Edunov, Myle Ott, Michael Auli, David Grangier, and Marc{'}Aurelio
  Ranzato. 2018.
\newblock \href {https://doi.org/10.18653/v1/N18-1033} {Classical structured
  prediction losses for sequence to sequence learning}.
\newblock In \emph{Proceedings of the 2018 Conference of the North {A}merican
  Chapter of the Association for Computational Linguistics: Human Language
  Technologies, Volume 1 (Long Papers)}, pages 355--364, New Orleans,
  Louisiana. Association for Computational Linguistics.

\bibitem[{Fu et~al.(2023)Fu, Ng, Jiang, and Liu}]{fu2023gptscore}
Jinlan Fu, See-Kiong Ng, Zhengbao Jiang, and Pengfei Liu. 2023.
\newblock Gptscore: Evaluate as you desire.
\newblock \emph{arXiv preprint arXiv:2302.04166}.

\bibitem[{Gao et~al.(2023)Gao, Ruan, Sun, Yin, Yang, and Wan}]{gao2023human}
Mingqi Gao, Jie Ruan, Renliang Sun, Xunjian Yin, Shiping Yang, and Xiaojun Wan.
  2023.
\newblock Human-like summarization evaluation with chatgpt.
\newblock \emph{arXiv preprint arXiv:2304.02554}.

\bibitem[{Geng et~al.(2023)Geng, Gudibande, Liu, Wallace, Abbeel, Levine, and
  Song}]{koala_blogpost_2023}
Xinyang Geng, Arnav Gudibande, Hao Liu, Eric Wallace, Pieter Abbeel, Sergey
  Levine, and Dawn Song. 2023.
\newblock \href {https://bair.berkeley.edu/blog/2023/04/03/koala/} {Koala: A
  dialogue model for academic research}.
\newblock Blog post.

\bibitem[{Geva et~al.(2021)Geva, Khashabi, Segal, Khot, Roth, and
  Berant}]{geva-etal-2021-aristotle}
Mor Geva, Daniel Khashabi, Elad Segal, Tushar Khot, Dan Roth, and Jonathan
  Berant. 2021.
\newblock \href {https://doi.org/10.1162/tacl_a_00370} {Did aristotle use a
  laptop? a question answering benchmark with implicit reasoning strategies}.
\newblock \emph{Transactions of the Association for Computational Linguistics},
  9:346--361.

\bibitem[{Ghosal et~al.(2023)Ghosal, Chia, Majumder, and
  Poria}]{ghosal2023flacuna}
Deepanway Ghosal, Yew~Ken Chia, Navonil Majumder, and Soujanya Poria. 2023.
\newblock Flacuna: Unleashing the problem solving power of vicuna using flan
  fine-tuning.
\newblock \emph{arXiv preprint arXiv:2307.02053}.

\bibitem[{Gunasekar et~al.(2023)Gunasekar, Zhang, Aneja, Mendes, Del~Giorno,
  Gopi, Javaheripi, Kauffmann, de~Rosa, Saarikivi
  et~al.}]{gunasekar2023textbooks}
Suriya Gunasekar, Yi~Zhang, Jyoti Aneja, Caio C{\'e}sar~Teodoro Mendes, Allie
  Del~Giorno, Sivakanth Gopi, Mojan Javaheripi, Piero Kauffmann, Gustavo
  de~Rosa, Olli Saarikivi, et~al. 2023.
\newblock Textbooks are all you need.
\newblock \emph{arXiv preprint arXiv:2306.11644}.

\bibitem[{He et~al.(2022)He, Zhou, Ma, Berg-Kirkpatrick, and
  Neubig}]{he2022towards}
Junxian He, Chunting Zhou, Xuezhe Ma, Taylor Berg-Kirkpatrick, and Graham
  Neubig. 2022.
\newblock \href {https://openreview.net/forum?id=0RDcd5Axok} {Towards a unified
  view of parameter-efficient transfer learning}.
\newblock In \emph{International Conference on Learning Representations}.

\bibitem[{Hendrycks et~al.(2021)Hendrycks, Burns, Basart, Zou, Mazeika, Song,
  and Steinhardt}]{hendrycks2021measuring}
Dan Hendrycks, Collin Burns, Steven Basart, Andy Zou, Mantas Mazeika, Dawn
  Song, and Jacob Steinhardt. 2021.
\newblock \href {https://openreview.net/forum?id=d7KBjmI3GmQ} {Measuring
  massive multitask language understanding}.
\newblock In \emph{International Conference on Learning Representations}.

\bibitem[{Honovich et~al.(2022)Honovich, Scialom, Levy, and
  Schick}]{DBLP:journals/corr/abs-2212-09689}
Or~Honovich, Thomas Scialom, Omer Levy, and Timo Schick. 2022.
\newblock \href {https://doi.org/10.48550/arXiv.2212.09689} {Unnatural
  instructions: Tuning language models with (almost) no human labor}.
\newblock \emph{CoRR}, abs/2212.09689.

\bibitem[{Hu et~al.(2022)Hu, yelong shen, Wallis, Allen-Zhu, Li, Wang, Wang,
  and Chen}]{hu2022lora}
Edward~J Hu, yelong shen, Phillip Wallis, Zeyuan Allen-Zhu, Yuanzhi Li, Shean
  Wang, Lu~Wang, and Weizhu Chen. 2022.
\newblock \href {https://openreview.net/forum?id=nZeVKeeFYf9} {Lo{RA}: Low-rank
  adaptation of large language models}.
\newblock In \emph{International Conference on Learning Representations}.

\bibitem[{Huang et~al.(2023)Huang, Bai, Zhu, Zhang, Zhang, Su, Liu, Lv, Zhang,
  Lei, Fu, Sun, and He}]{huang2023ceval}
Yuzhen Huang, Yuzhuo Bai, Zhihao Zhu, Junlei Zhang, Jinghan Zhang, Tangjun Su,
  Junteng Liu, Chuancheng Lv, Yikai Zhang, Jiayi Lei, Yao Fu, Maosong Sun, and
  Junxian He. 2023.
\newblock C-eval: A multi-level multi-discipline chinese evaluation suite for
  foundation models.
\newblock \emph{arXiv preprint arXiv:2305.08322}.

\bibitem[{Jentzsch and Kersting(2023)}]{jentzsch2023chatgpt}
Sophie Jentzsch and Kristian Kersting. 2023.
\newblock Chatgpt is fun, but it is not funny! humor is still challenging large
  language models.
\newblock \emph{arXiv preprint arXiv:2306.04563}.

\bibitem[{Ji et~al.(2023)Ji, Gong, Deng, Peng, Niu, Ma, and
  Li}]{DBLP:journals/corr/abs-2304-07854}
Yunjie Ji, Yan Gong, Yong Deng, Yiping Peng, Qiang Niu, Baochang Ma, and
  Xiangang Li. 2023.
\newblock \href {https://doi.org/10.48550/arXiv.2304.07854} {Towards better
  instruction following language models for chinese: Investigating the impact
  of training data and evaluation}.
\newblock \emph{CoRR}, abs/2304.07854.

\bibitem[{Jiang et~al.(2023)Jiang, Chan, Chen, and Wang}]{Jiang2023LionAD}
Yuxin Jiang, Chunkit Chan, Mingyang Chen, and Wei Wang. 2023.
\newblock Lion: Adversarial distillation of closed-source large language model.
\newblock \emph{ArXiv}, abs/2305.12870.

\bibitem[{Kalpathy-Cramer et~al.(2016)Kalpathy-Cramer, Campbell, Erdogmus,
  Tian, Kedarisetti, Moleta, Reynolds, Hutcheson, Shapiro, Repka, Ferrone,
  Drenser, Horowitz, Sonmez, Swan, Ostmo, Jonas, Chan, Chiang, Chiang, Ostmo,
  Sonmez, Campbell, Chan, Jonas, Horowitz, Coki, Eccles, Sarna, Berrocal,
  Negron, Denser, Cumming, Osentoski, Check, Zajechowski, Lee, Kruger,
  McGovern, Simmons, Murthy, Galvis, Rotter, Chen, Li, Taylor, Roll,
  Kalpathy-Cramer, Erdogmus, Martinez-Castellanos, Salinas-Longoria, Romero,
  Arriola, Olguin-Manriquez, Meraz-Gutierrez, Dulanto-Reinoso, and
  Montero-Mendoza}]{KALPATHYCRAMER20162345}
Jayashree Kalpathy-Cramer, J.~Peter Campbell, Deniz Erdogmus, Peng Tian,
  Dharanish Kedarisetti, Chace Moleta, James~D. Reynolds, Kelly Hutcheson,
  Michael~J. Shapiro, Michael~X. Repka, Philip Ferrone, Kimberly Drenser, Jason
  Horowitz, Kemal Sonmez, Ryan Swan, Susan Ostmo, Karyn~E. Jonas, R.V.~Paul
  Chan, Michael~F. Chiang, Michael~F. Chiang, Susan Ostmo, Kemal Sonmez,
  J.~Peter Campbell, R.V.~Paul Chan, Karyn Jonas, Jason Horowitz, Osode Coki,
  Cheryl-Ann Eccles, Leora Sarna, Audina Berrocal, Catherin Negron, Kimberly
  Denser, Kristi Cumming, Tammy Osentoski, Tammy Check, Mary Zajechowski,
  Thomas Lee, Evan Kruger, Kathryn McGovern, Charles Simmons, Raghu Murthy,
  Sharon Galvis, Jerome Rotter, Ida Chen, Xiaohui Li, Kent Taylor, Kaye Roll,
  Jayashree Kalpathy-Cramer, Deniz Erdogmus, Maria~Ana Martinez-Castellanos,
  Samantha Salinas-Longoria, Rafael Romero, Andrea Arriola, Francisco
  Olguin-Manriquez, Miroslava Meraz-Gutierrez, Carlos~M. Dulanto-Reinoso, and
  Cristina Montero-Mendoza. 2016.
\newblock \href {https://doi.org/https://doi.org/10.1016/j.ophtha.2016.07.020}
  {Plus disease in retinopathy of prematurity: Improving diagnosis by ranking
  disease severity and using quantitative image analysis}.
\newblock \emph{Ophthalmology}, 123(11):2345--2351.

\bibitem[{Kopf et~al.(2023)Kopf, Kilcher, von Rutte, Anagnostidis, Tam,
  Stevens, Barhoum, Duc, Stanley, Nagyfi, Shahul, Suri, Glushkov, Dantuluri,
  Maguire, Schuhmann, Nguyen, and Mattick}]{Kopf2023OpenAssistantC}
Andreas Kopf, Yannic Kilcher, Dimitri von Rutte, Sotiris Anagnostidis, Zhi~Rui
  Tam, Keith Stevens, Abdullah Barhoum, Nguyen~Minh Duc, Oliver Stanley,
  Rich'ard Nagyfi, ES~Shahul, Sameer Suri, David Glushkov, Arnav Dantuluri,
  Andrew Maguire, Christoph Schuhmann, Huu Nguyen, and Alexander Mattick. 2023.
\newblock Openassistant conversations - democratizing large language model
  alignment.
\newblock \emph{ArXiv}, abs/2304.07327.

\bibitem[{Lai et~al.(2022)Lai, Li, Wang, Zhang, Zhong, Zettlemoyer, tau Yih,
  Fried, Wang, and Yu}]{Lai2022DS1000}
Yuhang Lai, Chengxi Li, Yiming Wang, Tianyi Zhang, Ruiqi Zhong, Luke
  Zettlemoyer, Scott~Wen tau Yih, Daniel Fried, Sida Wang, and Tao Yu. 2022.
\newblock Ds-1000: A natural and reliable benchmark for data science code
  generation.
\newblock \emph{ArXiv}, abs/2211.11501.

\bibitem[{Lester et~al.(2021)Lester, Al-Rfou, and
  Constant}]{lester-etal-2021-power}
Brian Lester, Rami Al-Rfou, and Noah Constant. 2021.
\newblock \href {https://doi.org/10.18653/v1/2021.emnlp-main.243} {The power of
  scale for parameter-efficient prompt tuning}.
\newblock In \emph{Proceedings of the 2021 Conference on Empirical Methods in
  Natural Language Processing}, pages 3045--3059, Online and Punta Cana,
  Dominican Republic. Association for Computational Linguistics.

\bibitem[{Li et~al.(2023{\natexlab{a}})Li, Hammoud, Itani, Khizbullin, and
  Ghanem}]{DBLP:journals/corr/abs-2303-17760}
Guohao Li, Hasan Abed Al~Kader Hammoud, Hani Itani, Dmitrii Khizbullin, and
  Bernard Ghanem. 2023{\natexlab{a}}.
\newblock \href {https://doi.org/10.48550/arXiv.2303.17760} {{CAMEL:}
  communicative agents for "mind" exploration of large scale language model
  society}.
\newblock \emph{CoRR}, abs/2303.17760.

\bibitem[{Li et~al.(2023{\natexlab{b}})Li, Koto, Wu, Aji, and
  Baldwin}]{bactrian}
Haonan Li, Fajri Koto, Minghao Wu, Alham~Fikri Aji, and Timothy Baldwin.
  2023{\natexlab{b}}.
\newblock Bactrian-x: A multilingual replicable instruction-following model
  with low-rank adaptation.
\newblock \emph{arXiv preprint arXiv:2305.15011}.

\bibitem[{Li et~al.(2023{\natexlab{c}})Li, Zhang, Koto, Yang, Zhao, Gong, Duan,
  and Baldwin}]{li2023cmmlu}
Haonan Li, Yixuan Zhang, Fajri Koto, Yifei Yang, Hai Zhao, Yeyun Gong, Nan
  Duan, and Timothy Baldwin. 2023{\natexlab{c}}.
\newblock Cmmlu: Measuring massive multitask language understanding in chinese.
\newblock \emph{arXiv preprint arXiv:2306.09212}.

\bibitem[{Li and Liang(2021)}]{li-liang-2021-prefix}
Xiang~Lisa Li and Percy Liang. 2021.
\newblock \href {https://doi.org/10.18653/v1/2021.acl-long.353} {Prefix-tuning:
  Optimizing continuous prompts for generation}.
\newblock In \emph{Proceedings of the 59th Annual Meeting of the Association
  for Computational Linguistics and the 11th International Joint Conference on
  Natural Language Processing (Volume 1: Long Papers)}, pages 4582--4597,
  Online. Association for Computational Linguistics.

\bibitem[{Lin(2004)}]{lin-2004-rouge}
Chin-Yew Lin. 2004.
\newblock \href {https://aclanthology.org/W04-1013} {{ROUGE}: A package for
  automatic evaluation of summaries}.
\newblock In \emph{Text Summarization Branches Out}, pages 74--81, Barcelona,
  Spain. Association for Computational Linguistics.

\bibitem[{Lin and Chen(2023)}]{lin2023llm}
Yen-Ting Lin and Yun-Nung Chen. 2023.
\newblock Llm-eval: Unified multi-dimensional automatic evaluation for
  open-domain conversations with large language models.
\newblock \emph{arXiv preprint arXiv:2305.13711}.

\bibitem[{Liu et~al.(2023{\natexlab{a}})Liu, Jin, Ren, Yu, Dong, Peng, Zhang,
  Peng, Zhang, Lyu et~al.}]{liu2023m3ke}
Chuang Liu, Renren Jin, Yuqi Ren, Linhao Yu, Tianyu Dong, Xiaohan Peng, Shuting
  Zhang, Jianxiang Peng, Peiyi Zhang, Qingqing Lyu, et~al. 2023{\natexlab{a}}.
\newblock M3ke: A massive multi-level multi-subject knowledge evaluation
  benchmark for chinese large language models.
\newblock \emph{arXiv preprint arXiv:2305.10263}.

\bibitem[{Liu et~al.(2022{\natexlab{a}})Liu, Geng, Lee, Mordatch, Levine,
  Narang, and Abbeel}]{Liu2022TowardsBF}
Hao Liu, Xinyang Geng, Lisa Lee, Igor Mordatch, Sergey Levine, Sharan Narang,
  and P.~Abbeel. 2022{\natexlab{a}}.
\newblock Towards better few-shot and finetuning performance with forgetful
  causal language models.

\bibitem[{Liu et~al.(2023{\natexlab{b}})Liu, Sferrazza, and
  Abbeel}]{liu2023languages}
Hao Liu, Carmelo Sferrazza, and Pieter Abbeel. 2023{\natexlab{b}}.
\newblock Languages are rewards: Hindsight finetuning using human feedback.
\newblock \emph{arXiv preprint arXiv:2302.02676}.

\bibitem[{Liu et~al.(2023{\natexlab{c}})Liu, Xia, Wang, and
  Zhang}]{liu2023your}
Jiawei Liu, Chunqiu~Steven Xia, Yuyao Wang, and Lingming Zhang.
  2023{\natexlab{c}}.
\newblock Is your code generated by chatgpt really correct? rigorous evaluation
  of large language models for code generation.
\newblock \emph{arXiv preprint arXiv:2305.01210}.

\bibitem[{Liu et~al.(2022{\natexlab{b}})Liu, Jia, Zhang, Zhuang, Liu, and
  Vosoughi}]{liu2022second}
Ruibo Liu, Chenyan Jia, Ge~Zhang, Ziyu Zhuang, Tony~X Liu, and Soroush
  Vosoughi. 2022{\natexlab{b}}.
\newblock \href {https://openreview.net/forum?id=u6OfmaGIya1} {Second thoughts
  are best: Learning to re-align with human values from text edits}.
\newblock In \emph{Advances in Neural Information Processing Systems}.

\bibitem[{Liu et~al.(2023{\natexlab{d}})Liu, Yang, Jia, Zhang, Zhou, Dai, Yang,
  and Vosoughi}]{liu2023training}
Ruibo Liu, Ruixin Yang, Chenyan Jia, Ge~Zhang, Denny Zhou, Andrew~M Dai, Diyi
  Yang, and Soroush Vosoughi. 2023{\natexlab{d}}.
\newblock Training socially aligned language models in simulated human society.
\newblock \emph{arXiv preprint arXiv:2305.16960}.

\bibitem[{Liu et~al.(2023{\natexlab{e}})Liu, Iter, Xu, Wang, Xu, and
  Zhu}]{liu2023gpteval}
Yang Liu, Dan Iter, Yichong Xu, Shuohang Wang, Ruochen Xu, and Chenguang Zhu.
  2023{\natexlab{e}}.
\newblock Gpteval: Nlg evaluation using gpt-4 with better human alignment.
\newblock \emph{arXiv preprint arXiv:2303.16634}.

\bibitem[{Liu et~al.(2022{\natexlab{c}})Liu, Liu, Radev, and
  Neubig}]{liu-etal-2022-brio}
Yixin Liu, Pengfei Liu, Dragomir Radev, and Graham Neubig. 2022{\natexlab{c}}.
\newblock \href {https://doi.org/10.18653/v1/2022.acl-long.207} {{BRIO}:
  Bringing order to abstractive summarization}.
\newblock In \emph{Proceedings of the 60th Annual Meeting of the Association
  for Computational Linguistics (Volume 1: Long Papers)}, pages 2890--2903,
  Dublin, Ireland. Association for Computational Linguistics.

\bibitem[{Longpre et~al.(2023)Longpre, Hou, Vu, Webson, Chung, Tay, Zhou, Le,
  Zoph, Wei et~al.}]{longpre2023flan}
Shayne Longpre, Le~Hou, Tu~Vu, Albert Webson, Hyung~Won Chung, Yi~Tay, Denny
  Zhou, Quoc~V Le, Barret Zoph, Jason Wei, et~al. 2023.
\newblock The flan collection: Designing data and methods for effective
  instruction tuning.
\newblock \emph{arXiv preprint arXiv:2301.13688}.

\bibitem[{Luo et~al.(2023)Luo, Xu, Zhao, Sun, Geng, Hu, Tao, Ma, Lin, and
  Jiang}]{luo2023wizardcoder}
Ziyang Luo, Can Xu, Pu~Zhao, Qingfeng Sun, Xiubo Geng, Wenxiang Hu, Chongyang
  Tao, Jing Ma, Qingwei Lin, and Daxin Jiang. 2023.
\newblock Wizardcoder: Empowering code large language models with
  evol-instruct.
\newblock \emph{arXiv preprint arXiv:2306.08568}.

\bibitem[{Mangrulkar et~al.(2022)Mangrulkar, Gugger, Debut, Belkada, and
  Paul}]{peft}
Sourab Mangrulkar, Sylvain Gugger, Lysandre Debut, Younes Belkada, and Sayak
  Paul. 2022.
\newblock Peft: State-of-the-art parameter-efficient fine-tuning methods.
\newblock \url{https://github.com/huggingface/peft}.

\bibitem[{Min et~al.(2023)Min, Krishna, Lyu, Lewis, Yih, Koh, Iyyer,
  Zettlemoyer, and Hajishirzi}]{min2023factscore}
Sewon Min, Kalpesh Krishna, Xinxi Lyu, Mike Lewis, Wen-tau Yih, Pang~Wei Koh,
  Mohit Iyyer, Luke Zettlemoyer, and Hannaneh Hajishirzi. 2023.
\newblock Factscore: Fine-grained atomic evaluation of factual precision in
  long form text generation.
\newblock \emph{arXiv preprint arXiv:2305.14251}.

\bibitem[{Mishra et~al.(2022)Mishra, Khashabi, Baral, and
  Hajishirzi}]{mishra-etal-2022-cross}
Swaroop Mishra, Daniel Khashabi, Chitta Baral, and Hannaneh Hajishirzi. 2022.
\newblock \href {https://doi.org/10.18653/v1/2022.acl-long.244} {Cross-task
  generalization via natural language crowdsourcing instructions}.
\newblock In \emph{Proceedings of the 60th Annual Meeting of the Association
  for Computational Linguistics (Volume 1: Long Papers)}, pages 3470--3487,
  Dublin, Ireland. Association for Computational Linguistics.

\bibitem[{Muennighoff et~al.(2023)Muennighoff, Rush, Barak, Scao, Piktus, Tazi,
  Pyysalo, Wolf, and Raffel}]{muennighoff2023scaling}
Niklas Muennighoff, Alexander~M Rush, Boaz Barak, Teven~Le Scao, Aleksandra
  Piktus, Nouamane Tazi, Sampo Pyysalo, Thomas Wolf, and Colin Raffel. 2023.
\newblock Scaling data-constrained language models.
\newblock \emph{arXiv preprint arXiv:2305.16264}.

\bibitem[{Mukherjee et~al.(2023)Mukherjee, Mitra, Jawahar, Agarwal, Palangi,
  and Awadallah}]{mukherjee2023orca}
Subhabrata Mukherjee, Arindam Mitra, Ganesh Jawahar, Sahaj Agarwal, Hamid
  Palangi, and Ahmed Awadallah. 2023.
\newblock Orca: Progressive learning from complex explanation traces of gpt-4.
\newblock \emph{arXiv preprint arXiv:2306.02707}.

\bibitem[{Nguyen et~al.(2023)Nguyen, Suri, Tsui, and Schuhmann}]{OIG}
Huu Nguyen, Sameer Suri, Ken Tsui, and Christoph Schuhmann. 2023.
\newblock \href {https://laion.ai/blog/oig-dataset/} {The oig dataset}.

\bibitem[{Nguyen et~al.(2022)Nguyen, Zheng, and
  Grover}]{nguyen2022conserweightive}
Tung Nguyen, Qinqing Zheng, and Aditya Grover. 2022.
\newblock Conserweightive behavioral cloning for reliable offline reinforcement
  learning.
\newblock \emph{arXiv preprint arXiv:2210.05158}.

\bibitem[{Ouyang et~al.(2022)Ouyang, Wu, Jiang, Almeida, Wainwright, Mishkin,
  Zhang, Agarwal, Slama, Gray, Schulman, Hilton, Kelton, Miller, Simens,
  Askell, Welinder, Christiano, Leike, and
  Lowe}]{DBLP:conf/nips/Ouyang0JAWMZASR22}
Long Ouyang, Jeffrey Wu, Xu~Jiang, Diogo Almeida, Carroll Wainwright, Pamela
  Mishkin, Chong Zhang, Sandhini Agarwal, Katarina Slama, Alex Gray, John
  Schulman, Jacob Hilton, Fraser Kelton, Luke Miller, Maddie Simens, Amanda
  Askell, Peter Welinder, Paul Christiano, Jan Leike, and Ryan Lowe. 2022.
\newblock \href {https://openreview.net/forum?id=TG8KACxEON} {Training language
  models to follow instructions with human feedback}.
\newblock In \emph{Advances in Neural Information Processing Systems}.

\bibitem[{Papineni et~al.(2002)Papineni, Roukos, Ward, and
  Zhu}]{papineni-etal-2002-bleu}
Kishore Papineni, Salim Roukos, Todd Ward, and Wei-Jing Zhu. 2002.
\newblock \href {https://doi.org/10.3115/1073083.1073135} {{B}leu: a method for
  automatic evaluation of machine translation}.
\newblock In \emph{Proceedings of the 40th Annual Meeting of the Association
  for Computational Linguistics}, pages 311--318, Philadelphia, Pennsylvania,
  USA. Association for Computational Linguistics.

\bibitem[{Peng et~al.(2023)Peng, Li, He, Galley, and Gao}]{peng2023instruction}
Baolin Peng, Chunyuan Li, Pengcheng He, Michel Galley, and Jianfeng Gao. 2023.
\newblock Instruction tuning with gpt-4.
\newblock \emph{arXiv preprint arXiv:2304.03277}.

\bibitem[{Rafailov et~al.(2023)Rafailov, Sharma, Mitchell, Ermon, Manning, and
  Finn}]{rafailov2023direct}
Rafael Rafailov, Archit Sharma, Eric Mitchell, Stefano Ermon, Christopher~D
  Manning, and Chelsea Finn. 2023.
\newblock Direct preference optimization: Your language model is secretly a
  reward model.
\newblock \emph{arXiv preprint arXiv:2305.18290}.

\bibitem[{Song et~al.(2023)Song, Yu, Li, Yu, Huang, Li, and
  Wang}]{song2023preference}
Feifan Song, Bowen Yu, Minghao Li, Haiyang Yu, Fei Huang, Yongbin Li, and
  Houfeng Wang. 2023.
\newblock Preference ranking optimization for human alignment.
\newblock \emph{arXiv preprint arXiv:2306.17492}.

\bibitem[{Sun et~al.(2023{\natexlab{a}})Sun, Ji, Ma, and
  Li}]{sun2023comparative}
Xianghui Sun, Yunjie Ji, Baochang Ma, and Xiangang Li. 2023{\natexlab{a}}.
\newblock A comparative study between full-parameter and lora-based fine-tuning
  on chinese instruction data for instruction following large language model.
\newblock \emph{arXiv preprint arXiv:2304.08109}.

\bibitem[{Sun et~al.(2023{\natexlab{b}})Sun, Shen, Zhou, Zhang, Chen, Cox,
  Yang, and Gan}]{Sun2023PrincipleDrivenSO}
Zhiqing Sun, Yikang Shen, Qinhong Zhou, Hongxin Zhang, Zhenfang Chen, David~D.
  Cox, Yiming Yang, and Chuang Gan. 2023{\natexlab{b}}.
\newblock Principle-driven self-alignment of language models from scratch with
  minimal human supervision.

\bibitem[{Suzgun et~al.(2022)Suzgun, Scales, Sch{\"a}rli, Gehrmann, Tay, Chung,
  Chowdhery, Le, Chi, Zhou, , and Wei}]{suzgun2022challenging}
Mirac Suzgun, Nathan Scales, Nathanael Sch{\"a}rli, Sebastian Gehrmann, Yi~Tay,
  Hyung~Won Chung, Aakanksha Chowdhery, Quoc~V Le, Ed~H Chi, Denny Zhou, , and
  Jason Wei. 2022.
\newblock Challenging big-bench tasks and whether chain-of-thought can solve
  them.
\newblock \emph{arXiv preprint arXiv:2210.09261}.

\bibitem[{Talmor et~al.(2019)Talmor, Herzig, Lourie, and
  Berant}]{talmor-etal-2019-commonsenseqa}
Alon Talmor, Jonathan Herzig, Nicholas Lourie, and Jonathan Berant. 2019.
\newblock \href {https://doi.org/10.18653/v1/N19-1421} {{C}ommonsense{QA}: A
  question answering challenge targeting commonsense knowledge}.
\newblock In \emph{Proceedings of the 2019 Conference of the North {A}merican
  Chapter of the Association for Computational Linguistics: Human Language
  Technologies, Volume 1 (Long and Short Papers)}, pages 4149--4158,
  Minneapolis, Minnesota. Association for Computational Linguistics.

\bibitem[{Tang et~al.(2023)Tang, Lu, Jiang, Huang, Zhang, Zhao, and
  Wei}]{tang2023not}
Tianyi Tang, Hongyuan Lu, Yuchen~Eleanor Jiang, Haoyang Huang, Dongdong Zhang,
  Wayne~Xin Zhao, and Furu Wei. 2023.
\newblock Not all metrics are guilty: Improving nlg evaluation with llm
  paraphrasing.
\newblock \emph{arXiv preprint arXiv:2305.15067}.

\bibitem[{Taori et~al.(2023)Taori, Gulrajani, Zhang, Dubois, Li, Guestrin,
  Liang, and Hashimoto}]{alpaca}
Rohan Taori, Ishaan Gulrajani, Tianyi Zhang, Yann Dubois, Xuechen Li, Carlos
  Guestrin, Percy Liang, and Tatsunori~B. Hashimoto. 2023.
\newblock Stanford alpaca: An instruction-following llama model.
\newblock \url{https://github.com/tatsu-lab/stanford_alpaca}.

\bibitem[{Team(2023)}]{MosaicML2023Introducing}
MosaicML~NLP Team. 2023.
\newblock \href {www.mosaicml.com/blog/mpt-30b} {Introducing mpt-30b: Raising
  the bar for open-source foundation models}.
\newblock Accessed: 2023-06-22.

\bibitem[{Wang et~al.(2023{\natexlab{a}})Wang, Cheng, Yu, and Liu}]{openchat}
Guan Wang, Sijie Cheng, Qiying Yu, and Changling Liu. 2023{\natexlab{a}}.
\newblock \href {https://doi.org/10.5281/zenodo.8105775} {{OpenChat: Advancing
  Open-source Language Models with Imperfect Data}}.

\bibitem[{Wang et~al.(2023{\natexlab{b}})Wang, Li, Chen, Zhu, Lin, Cao, Liu,
  Liu, and Sui}]{wang2023large}
Peiyi Wang, Lei Li, Liang Chen, Dawei Zhu, Binghuai Lin, Yunbo Cao, Qi~Liu,
  Tianyu Liu, and Zhifang Sui. 2023{\natexlab{b}}.
\newblock Large language models are not fair evaluators.
\newblock \emph{arXiv preprint arXiv:2305.17926}.

\bibitem[{Wang et~al.(2023{\natexlab{c}})Wang, Yu, Zeng, Yang, Wang, Chen,
  Jiang, Xie, Wang, Xie et~al.}]{wang2023pandalm}
Yidong Wang, Zhuohao Yu, Zhengran Zeng, Linyi Yang, Cunxiang Wang, Hao Chen,
  Chaoya Jiang, Rui Xie, Jindong Wang, Xing Xie, et~al. 2023{\natexlab{c}}.
\newblock Pandalm: An automatic evaluation benchmark for llm instruction tuning
  optimization.
\newblock \emph{arXiv preprint arXiv:2306.05087}.

\bibitem[{Wang et~al.(2023{\natexlab{d}})Wang, Ivison, Dasigi, Hessel, Khot,
  Chandu, Wadden, MacMillan, Smith, Beltagy et~al.}]{wang2023far}
Yizhong Wang, Hamish Ivison, Pradeep Dasigi, Jack Hessel, Tushar Khot,
  Khyathi~Raghavi Chandu, David Wadden, Kelsey MacMillan, Noah~A Smith,
  Iz~Beltagy, et~al. 2023{\natexlab{d}}.
\newblock How far can camels go? exploring the state of instruction tuning on
  open resources.
\newblock \emph{arXiv preprint arXiv:2306.04751}.

\bibitem[{Wang et~al.(2022{\natexlab{a}})Wang, Kordi, Mishra, Liu, Smith,
  Khashabi, and Hajishirzi}]{DBLP:journals/corr/abs-2212-10560}
Yizhong Wang, Yeganeh Kordi, Swaroop Mishra, Alisa Liu, Noah~A. Smith, Daniel
  Khashabi, and Hannaneh Hajishirzi. 2022{\natexlab{a}}.
\newblock \href {https://doi.org/10.48550/arXiv.2212.10560} {Self-instruct:
  Aligning language model with self generated instructions}.
\newblock \emph{CoRR}, abs/2212.10560.

\bibitem[{Wang et~al.(2022{\natexlab{b}})Wang, Mishra, Alipoormolabashi, Kordi,
  Mirzaei, Naik, Ashok, Dhanasekaran, Arunkumar, Stap, Pathak, Karamanolakis,
  Lai, Purohit, Mondal, Anderson, Kuznia, Doshi, Pal, Patel, Moradshahi,
  Parmar, Purohit, Varshney, Kaza, Verma, Puri, Karia, Doshi, Sampat, Mishra,
  Reddy~A, Patro, Dixit, and Shen}]{wang-etal-2022-super}
Yizhong Wang, Swaroop Mishra, Pegah Alipoormolabashi, Yeganeh Kordi, Amirreza
  Mirzaei, Atharva Naik, Arjun Ashok, Arut~Selvan Dhanasekaran, Anjana
  Arunkumar, David Stap, Eshaan Pathak, Giannis Karamanolakis, Haizhi Lai,
  Ishan Purohit, Ishani Mondal, Jacob Anderson, Kirby Kuznia, Krima Doshi,
  Kuntal~Kumar Pal, Maitreya Patel, Mehrad Moradshahi, Mihir Parmar, Mirali
  Purohit, Neeraj Varshney, Phani~Rohitha Kaza, Pulkit Verma, Ravsehaj~Singh
  Puri, Rushang Karia, Savan Doshi, Shailaja~Keyur Sampat, Siddhartha Mishra,
  Sujan Reddy~A, Sumanta Patro, Tanay Dixit, and Xudong Shen.
  2022{\natexlab{b}}.
\newblock \href {https://aclanthology.org/2022.emnlp-main.340}
  {Super-{N}atural{I}nstructions: Generalization via declarative instructions
  on 1600+ {NLP} tasks}.
\newblock In \emph{Proceedings of the 2022 Conference on Empirical Methods in
  Natural Language Processing}, pages 5085--5109, Abu Dhabi, United Arab
  Emirates. Association for Computational Linguistics.

\bibitem[{Wei et~al.(2022{\natexlab{a}})Wei, Bosma, Zhao, Guu, Yu, Lester, Du,
  Dai, and Le}]{wei2022finetuned}
Jason Wei, Maarten Bosma, Vincent Zhao, Kelvin Guu, Adams~Wei Yu, Brian Lester,
  Nan Du, Andrew~M. Dai, and Quoc~V Le. 2022{\natexlab{a}}.
\newblock \href {https://openreview.net/forum?id=gEZrGCozdqR} {Finetuned
  language models are zero-shot learners}.
\newblock In \emph{International Conference on Learning Representations}.

\bibitem[{Wei et~al.(2022{\natexlab{b}})Wei, Wang, Schuurmans, Bosma, brian
  ichter, Xia, Chi, Le, and Zhou}]{wei2022chain}
Jason Wei, Xuezhi Wang, Dale Schuurmans, Maarten Bosma, brian ichter, Fei Xia,
  Ed~H. Chi, Quoc~V Le, and Denny Zhou. 2022{\natexlab{b}}.
\newblock \href {https://openreview.net/forum?id=_VjQlMeSB_J} {Chain of thought
  prompting elicits reasoning in large language models}.
\newblock In \emph{Advances in Neural Information Processing Systems}.

\bibitem[{Wu and Aji(2023)}]{Wu2023StyleOS}
Minghao Wu and Alham~Fikri Aji. 2023.
\newblock Style over substance: Evaluation biases for large language models.
\newblock \emph{ArXiv}, abs/2307.03025.

\bibitem[{Wu et~al.(2023)Wu, Waheed, Zhang, Abdul{-}Mageed, and
  Aji}]{DBLP:journals/corr/abs-2304-1440}
Minghao Wu, Abdul Waheed, Chiyu Zhang, Muhammad Abdul{-}Mageed, and Alham~Fikri
  Aji. 2023.
\newblock \href {https://doi.org/10.48550/arXiv.2304.14402} {Lamini-lm: {A}
  diverse herd of distilled models from large-scale instructions}.
\newblock \emph{CoRR}, abs/2304.14402.

\bibitem[{Xu et~al.(2023{\natexlab{a}})Xu, Yang, Lin, Wang, Zhou, Zhang, and
  Mao}]{xu2023expertprompting}
Benfeng Xu, An~Yang, Junyang Lin, Quan Wang, Chang Zhou, Yongdong Zhang, and
  Zhendong Mao. 2023{\natexlab{a}}.
\newblock Expertprompting: Instructing large language models to be
  distinguished experts.
\newblock \emph{arXiv preprint arXiv:2305.14688}.

\bibitem[{Xu et~al.(2023{\natexlab{b}})Xu, Sun, Zheng, Geng, Zhao, Feng, Tao,
  and Jiang}]{xu2023wizardlm}
Can Xu, Qingfeng Sun, Kai Zheng, Xiubo Geng, Pu~Zhao, Jiazhan Feng, Chongyang
  Tao, and Daxin Jiang. 2023{\natexlab{b}}.
\newblock \href {http://arxiv.org/abs/2304.12244} {Wizardlm: Empowering large
  language models to follow complex instructions}.

\bibitem[{Xu et~al.(2023{\natexlab{c}})Xu, Guo, Duan, and
  McAuley}]{DBLP:journals/corr/abs-2304-01196}
Canwen Xu, Daya Guo, Nan Duan, and Julian~J. McAuley. 2023{\natexlab{c}}.
\newblock \href {https://doi.org/10.48550/arXiv.2304.01196} {Baize: An
  open-source chat model with parameter-efficient tuning on self-chat data}.
\newblock \emph{CoRR}, abs/2304.01196.

\bibitem[{Ye et~al.(2023{\natexlab{a}})Ye, Jo, Kim, Kim, Hwang, and
  Seo}]{selfee2023}
Seonghyeon Ye, Yongrae Jo, Doyoung Kim, Sungdong Kim, Hyeonbin Hwang, and
  Minjoon Seo. 2023{\natexlab{a}}.
\newblock \href {https://kaistai.github.io/SelFee/} {Selfee: Iterative
  self-revising llm empowered by self-feedback generation}.
\newblock Blog post.

\bibitem[{Ye et~al.(2023{\natexlab{b}})Ye, Kim, Kim, Hwang, Kim, Jo, Thorne,
  Kim, and Seo}]{Ye2023FLASKFL}
Seonghyeon Ye, Doyoung Kim, Sungdong Kim, Hyeonbin Hwang, Seungone Kim, Yongrae
  Jo, James Thorne, Juho Kim, and Minjoon Seo. 2023{\natexlab{b}}.
\newblock Flask: Fine-grained language model evaluation based on alignment
  skill sets.

\bibitem[{Yu et~al.(2023{\natexlab{a}})Yu, Wang, Tu, Cao, Zhang-Li, Lv, Peng,
  Yao, Zhang, Li et~al.}]{yu2023kola}
Jifan Yu, Xiaozhi Wang, Shangqing Tu, Shulin Cao, Daniel Zhang-Li, Xin Lv, Hao
  Peng, Zijun Yao, Xiaohan Zhang, Hanming Li, et~al. 2023{\natexlab{a}}.
\newblock Kola: Carefully benchmarking world knowledge of large language
  models.
\newblock \emph{arXiv preprint arXiv:2306.09296}.

\bibitem[{Yu et~al.(2023{\natexlab{b}})Yu, Zhuang, Zhang, Meng, Ratner,
  Krishna, Shen, and Zhang}]{yu2023large}
Yue Yu, Yuchen Zhuang, Jieyu Zhang, Yu~Meng, Alexander Ratner, Ranjay Krishna,
  Jiaming Shen, and Chao Zhang. 2023{\natexlab{b}}.
\newblock Large language model as attributed training data generator: A tale of
  diversity and bias.
\newblock \emph{arXiv preprint arXiv:2306.15895}.

\bibitem[{Yuan et~al.(2023)Yuan, Yuan, Tan, Wang, Huang, and
  Huang}]{yuan2023rrhf}
Zheng Yuan, Hongyi Yuan, Chuanqi Tan, Wei Wang, Songfang Huang, and Fei Huang.
  2023.
\newblock \href {http://arxiv.org/abs/2304.05302} {Rrhf: Rank responses to
  align language models with human feedback without tears}.

\bibitem[{Zha et~al.(2023)Zha, Yang, Li, and Hu}]{zha2023alignscore}
Yuheng Zha, Yichi Yang, Ruichen Li, and Zhiting Hu. 2023.
\newblock Alignscore: Evaluating factual consistency with a unified alignment
  function.
\newblock \emph{arXiv preprint arXiv:2305.16739}.

\bibitem[{Zhang et~al.(2023{\natexlab{a}})Zhang, Shi, Liu, Yuan, Li, Dong, Shu,
  Li, Wang, Lin, Huang, and Fu}]{Zhang2023ChineseOI}
Ge~Zhang, Yemin Shi, Ruibo Liu, Ruibin Yuan, Yizhi Li, Siwei Dong, Yu~Shu,
  Zhaoqun Li, Zekun Wang, Chenghua Lin, Wen-Fen Huang, and Jie Fu.
  2023{\natexlab{a}}.
\newblock Chinese open instruction generalist: A preliminary release.
\newblock \emph{ArXiv}, abs/2304.07987.

\bibitem[{Zhang et~al.(2023{\natexlab{b}})Zhang, Chen, Bukharin, He, Cheng,
  Chen, and Zhao}]{zhang2023adaptive}
Qingru Zhang, Minshuo Chen, Alexander Bukharin, Pengcheng He, Yu~Cheng, Weizhu
  Chen, and Tuo Zhao. 2023{\natexlab{b}}.
\newblock \href {https://openreview.net/forum?id=lq62uWRJjiY} {Adaptive budget
  allocation for parameter-efficient fine-tuning}.
\newblock In \emph{The Eleventh International Conference on Learning
  Representations}.

\bibitem[{Zhang et~al.(2023{\natexlab{c}})Zhang, Fang, Zhang, Ma, Zhou, Huang,
  Bu, Gui, Chen, Chen, and Feng}]{Zhang2023BayLingBC}
Shaolei Zhang, Qingkai Fang, Zhuocheng Zhang, Zhengrui Ma, Yan Zhou, Langlin
  Huang, Mengyu Bu, Shangtong Gui, Yunji Chen, Xilin Chen, and Yang Feng.
  2023{\natexlab{c}}.
\newblock Bayling: Bridging cross-lingual alignment and instruction following
  through interactive translation for large language models.
\newblock \emph{ArXiv}, abs/2306.10968.

\bibitem[{Zhang* et~al.(2020)Zhang*, Kishore*, Wu*, Weinberger, and
  Artzi}]{Zhang*2020BERTScore:}
Tianyi Zhang*, Varsha Kishore*, Felix Wu*, Kilian~Q. Weinberger, and Yoav
  Artzi. 2020.
\newblock \href {https://openreview.net/forum?id=SkeHuCVFDr} {Bertscore:
  Evaluating text generation with bert}.
\newblock In \emph{International Conference on Learning Representations}.

\bibitem[{Zhao et~al.(2023)Zhao, Khalman, Joshi, Narayan, Saleh, and
  Liu}]{zhao2023calibrating}
Yao Zhao, Mikhail Khalman, Rishabh Joshi, Shashi Narayan, Mohammad Saleh, and
  Peter~J Liu. 2023.
\newblock \href {https://openreview.net/forum?id=0qSOodKmJaN} {Calibrating
  sequence likelihood improves conditional language generation}.
\newblock In \emph{The Eleventh International Conference on Learning
  Representations}.

\bibitem[{Zheng et~al.(2023)Zheng, Chiang, Sheng, Zhuang, Wu, Zhuang, Lin, Li,
  Li, Xing et~al.}]{zheng2023judging}
Lianmin Zheng, Wei-Lin Chiang, Ying Sheng, Siyuan Zhuang, Zhanghao Wu, Yonghao
  Zhuang, Zi~Lin, Zhuohan Li, Dacheng Li, Eric Xing, et~al. 2023.
\newblock Judging llm-as-a-judge with mt-bench and chatbot arena.
\newblock \emph{arXiv preprint arXiv:2306.05685}.

\bibitem[{Zhong et~al.(2023)Zhong, Cui, Guo, Liang, Lu, Wang, Saied, Chen, and
  Duan}]{zhong2023agieval}
Wanjun Zhong, Ruixiang Cui, Yiduo Guo, Yaobo Liang, Shuai Lu, Yanlin Wang, Amin
  Saied, Weizhu Chen, and Nan Duan. 2023.
\newblock Agieval: A human-centric benchmark for evaluating foundation models.
\newblock \emph{arXiv preprint arXiv:2304.06364}.

\bibitem[{Zhou et~al.(2023)Zhou, Liu, Xu, Iyer, Sun, Mao, Ma, Efrat, Yu, Yu
  et~al.}]{zhou2023lima}
Chunting Zhou, Pengfei Liu, Puxin Xu, Srini Iyer, Jiao Sun, Yuning Mao, Xuezhe
  Ma, Avia Efrat, Ping Yu, Lili Yu, et~al. 2023.
\newblock Lima: Less is more for alignment.
\newblock \emph{arXiv preprint arXiv:2305.11206}.

\bibitem[{Zhuo(2023)}]{zhuo2023large}
Terry~Yue Zhuo. 2023.
\newblock Large language models are state-of-the-art evaluators of code
  generation.
\newblock \emph{arXiv preprint arXiv:2304.14317}.

\bibitem[{Ziegler et~al.(2019)Ziegler, Stiennon, Wu, Brown, Radford, Amodei,
  Christiano, and Irving}]{ziegler2019finetuning}
Daniel~M. Ziegler, Nisan Stiennon, Jeffrey Wu, Tom~B. Brown, Alec Radford,
  Dario Amodei, Paul Christiano, and Geoffrey Irving. 2019.
\newblock \href {https://arxiv.org/abs/1909.08593} {Fine-tuning language models
  from human preferences}.
\newblock \emph{arXiv preprint arXiv:1909.08593}.

\end{thebibliography}
\bibliographystyle{acl_natbib}

\appendix

\section{Appendix}
\label{sec:appendix}
\begin{CJK*}{UTF8}{gkai}
\begin{table}[h]
\small
\caption{\label{tokenizer-comparison} The outputs of original LLaMA and Chinese Tokenizer. This example is from~\citet{cui2023efficient}.}
\begin{center}
\begin{tabular}{l}
\toprule
\makecell[lt]{\textbf{Inputs:} 人工智能是计算机科学、心理学、哲学等\\学科融合的交叉学科。} \\
\midrule
\makecell[lt]{\textbf{LLaMA:} \_, 人, 工, 智, 能, 是, 计, 算, 机, 科, 学, 、, 心, \\ 理, 学, 、, 0xE5, 0x93, 0xB2, 学, 等, 学, 科, 0xE8, \\ 0x9E, 0x8D, 合, 的, 交, 0xE5, 0x8F, 0x89, 学, 科, 。} \\
\midrule
\makecell[lt]{\textbf{Chinese:} \_, 人工智能, 是, 计算机, 科学, 、, 心理学, 、\\, 哲学, 等, 学科, 融合, 的, 交叉, 学科, 。} \\
\bottomrule
\end{tabular}
\end{center}
\end{table}
\end{CJK*}

\subsection{Training Language-Specific LLMs}
\label{otherlanguageLLMs}
Existing LLMs described above are mostly English-oriented. Thus, it becomes necessary to adapt the superior linguistic ability to other languages.~\citet{DBLP:journals/corr/abs-2304-07854,cui2023efficient} demonstrate existing English-dominated LLaMA has less than 1,000 Chinese characters in its vocabulary and LLaMA has to represent Chinese characters using the byte-based fallback strategy, which significantly increases input length and decreases the inference efficiency. As shown in Table~\ref{tokenizer-comparison}, compared to the default LLaMA tokenizer, the specialized Chinese tokenizer trained using large-scale Chinese corpus can produce more compact and semantically meaningful token representations (e.g., long and complex Chinese phrases). To leverage the linguistic knowledge in orginal LLaMA, ~\citet{cui2023efficient} propose a two-stage Chinese pre-training solution to enable LLaMA to better understand Chinese inputs. Before training they first add 20K Chinese words and phrases into the existing LLaMA vocabulary. In the first stage, they only train the input word embeddings and keep the rest parameters in LLaMA frozen. In the second stage, to save training resources, they add LoRA parameters and jointly train the parameters in the input word embeddings, self-attentive heads and LoRA parameters.~\citet{DBLP:journals/corr/abs-2304-07854} also report the benefits of such strategy under a GPT-4 evaluation framework.

\end{document}